\documentclass[conference]{IEEEtran}
\usepackage{times}

\usepackage[numbers]{natbib}
\usepackage{multicol}
\usepackage{flushend}
\usepackage[hidelinks]{hyperref}
\usepackage{url}

\usepackage{xcolor}
\usepackage{graphicx}
\usepackage{amsmath}
\usepackage{booktabs}
\usepackage{algorithm}

\usepackage{algpseudocode}
\usepackage{amsthm}
\usepackage{amssymb}
\usepackage{multirow}
\usepackage{tablefootnote}

\usepackage{mathtools}
\usepackage{booktabs}
\usepackage{amsfonts}
\usepackage{bbm}
\usepackage{caption}
\usepackage{subcaption}
\usepackage{amsmath}
\usepackage{amsfonts}

\usepackage{graphics}
\usepackage{bm}
\usepackage{float}
\usepackage{makecell}
\usepackage{wrapfig}
\usepackage{mathrsfs} 

\newcommand*\diff{\mathop{}\!\mathrm{d}}

\makeatletter
\newcommand*{\transpose}{%
  {\mathpalette\@transpose{}}%
}
\newcommand*{\@transpose}[2]{%
  \raisebox{\depth}{$\m@th#1\intercal$}%
}
\makeatother

\usepackage{xparse}
\usepackage[normalem]{ulem}

\usepackage{amssymb}

\NewDocumentCommand\NewIndexedVar{mmm}{
\expandafter\NewDocumentCommand\csname#1\endcsname{se{^_}}{#2\IfNoValueTF{##2}{}{^{##2}}\IfNoValueTF{##3}{\IfBooleanTF{##1}{}{_#3}}{_{##3}}}
}

\usepackage[acronym]{glossaries}
\newacronym{lfp}{LfP}{Learning from Play}
\newacronym{mtil}{MTIL}{Multi-Task Imitation Learning}
\newacronym{mlp}{MLP}{Multi-Layer Perceptron}
\newacronym{mdt}{MDT}{Multimodal Diffusion Transformer}
\newacronym{mtact}{MT-ACT}{Multi-Task Action Chunking Transformer}
\newacronym{mgf}{MGF}{Masked Generative Foresight}
\newacronym{vae}{VAE}{Variational Autoencoder}
\newacronym{mse}{MSE}{Mean Squared Error}
\newacronym{vit}{ViT}{Vision Transformer}
\newacronym{cla}{CLA}{Contrastive Latent Alignment}

\NewDocumentCommand\Dset{}{\mathcal{D}}
\NewDocumentCommand\Tset{}{\mathcal{T}}

\NewDocumentCommand\seq{}{\boldsymbol{o}}
\NewDocumentCommand\goal{}{\boldsymbol{g}}
\NewDocumentCommand\goalSet{}{\mathcal{G}}
\NewDocumentCommand\goalI{}{\textbf{o}}
\NewDocumentCommand\goalL{}{\textbf{l}}
\NewDocumentCommand\patch{}{\textbf{u}}
\NewDocumentCommand\clavar{}{\boldsymbol{z}}

\NewDocumentCommand\clatemp{}{\upsilon}
\NewDocumentCommand\cossim{}{C}
\NewDocumentCommand\Npatch{}{U}
\NewDocumentCommand\maskpercent{m}{U}
\NewDocumentCommand\p{mm}{#1\left(#2\right)}

\NewDocumentCommand\act{s O{i}}{\boldsymbol{\IfBooleanTF{#1}{a}{\bar{a}}}_{#2}}
\NewDocumentCommand\state{O{i}}{\boldsymbol{s}_{#1}}

\begin{document}

\title{Multimodal Diffusion Transformer: Learning Versatile Behavior from Multimodal Goals}


\author{\authorblockN{Moritz Reuss,  Ömer Erdinç Yağmurlu, Fabian Wenzel, Rudolf Lioutikov}
\authorblockA{Intuitive Robots Lab, 
Karlsruhe Institute of Technology, Germany}
}


%

\maketitle

\begin{abstract}
This work introduces the Multimodal Diffusion Transformer (MDT), a novel diffusion policy framework, that excels at learning versatile behavior from multimodal goal specifications with few language annotations.
MDT leverages a diffusion-based multimodal transformer backbone and two self-supervised auxiliary objectives to master long-horizon manipulation tasks based on multimodal goals.
The vast majority of imitation learning methods only learn from individual goal modalities, e.g. either language or goal images.
However, existing large-scale imitation learning datasets are only partially labeled with language annotations, which prohibits current methods from learning language conditioned behavior from these datasets.
MDT addresses this challenge by introducing a latent goal-conditioned state representation that is simultaneously trained on multimodal goal instructions. 
This state representation aligns image and language based goal embeddings and encodes sufficient information to predict future states.
The representation is trained via two self-supervised auxiliary objectives, enhancing the performance of the presented transformer backbone.
MDT shows exceptional performance on 164 tasks provided by the challenging CALVIN and LIBERO benchmarks, including a LIBERO version that contains less than $2\%$ language annotations.
Furthermore, MDT establishes a new record on the CALVIN manipulation challenge, demonstrating an absolute performance improvement of $15\%$ over prior state-of-the-art methods that require large-scale pretraining and contain $10\times$ more learnable parameters. 
MDT shows its ability to solve long-horizon manipulation from sparsely annotated data in both simulated and real-world environments. 
Demonstrations and Code are available at \url{https://intuitive-robots.github.io/mdt_policy/}.

\end{abstract}

\IEEEpeerreviewmaketitle

\section{Introduction}

Future robot agents need the ability to exhibit desired behavior according to intuitive instructions, similar to how humans interpret language or visual cues to understand tasks.
Current methods, however, often limit agents to process either language instructions~\cite{shridhar2022peract, wu2023unleashing, shridhar2023perceiver} or visual goals~\cite{cui2023from, reuss2023goal}.
This restriction limits the scope of training to fully-labeled datasets, which is not scalable for creating versatile robotic agents.

Natural language commands offer the greatest flexibility to instruct robots, as it is an intuitive form of communication for humans and it has become the most popular conditioning method for robots in recent years~\cite{shridhar2022peract, shridhar2023perceiver, zitkovich2023rt}.
However, training robots based on language instructions remains a significant challenge. 
\gls{mtil} has emerged as a promising approach, teaching robot agents a wide range of skills via learning from diverse human demonstrations~\cite{lynch2020language, lynch2023interactive}.
Unfortunately, \gls{mtil} capitalizes on large, fully annotated datasets. 
Collecting real human demonstrations is notably time-consuming and labor-intensive.

One way to circumvent these challenges is \gls{lfp}~\cite{lynch2020learning, mees2022calvin}, which capitalizes on large uncurated datasets.
\gls{lfp} allows for the fast collection of diverse demonstrations since it does not depend on scene staging, task segmentation, or resetting experiments~\cite{lynch2020learning}. 
Since these datasets are collected in such an uncurated way, they usually contain very few language annotations.
However, most current \gls{mtil} methods require language annotations for their entire training set, leaving these methods with too few demonstrations to train effective policies.
In contrast, future \gls{mtil} methods should be able to efficiently utilize the potential of diverse, cross-embodiment datasets like Open-X-Embodiment~\cite{open_x_embodiment_rt_x_2023}, with sparse language annotations.
This work introduces a novel approach that efficiently learns from multimodal goals, and hence efficiently leverages datasets with sparse language annotations.

Recently, Diffusion Generative Models have emerged as an effective policy representation for robot learning~\cite{chi2023diffusionpolicy, reuss2023goal}.
Diffusion Policies can learn expressive, versatile behavior conditioned on language-goals~\cite{xian2023unifying, ha2023scaling}.
Yet, none of the current methods adequately addresses the challenge of learning from multimodal goal specifications. 

This work introduces a novel diffusion-based approach able to learn versatile behavior from different goal modalities, such as language and images, simultaneously.
The approach learns efficiently even when trained on data with few language-annotated demonstrations.
The performance is further improved by introducing two simple, yet highly effective self-supervised losses, \gls{mgf} and \gls{cla}.
These losses encourage policies to learn latent features, that encode sufficient information to reconstruct partially-masked future frames conditioned on multimodal goals.
Hence, \gls{mgf} leverages the insight that policies benefit from informative latent spaces, which map goals to desired future states independent of their modality.
Detailed experiments and ablations show that the additional losses enhances the performance of current state-of-the-art transformer and diffusion policies, with minimal computational overhead.
The introduced \gls{mdt} approach combines the strengths of multimodal transformers with \gls{mgf} and latent token alignment. 
\gls{mdt} learns versatile behavior capable of following instructions provided as language or image goals. 

\begin{figure*}
    \centering
    \includegraphics[width=0.95\textwidth]{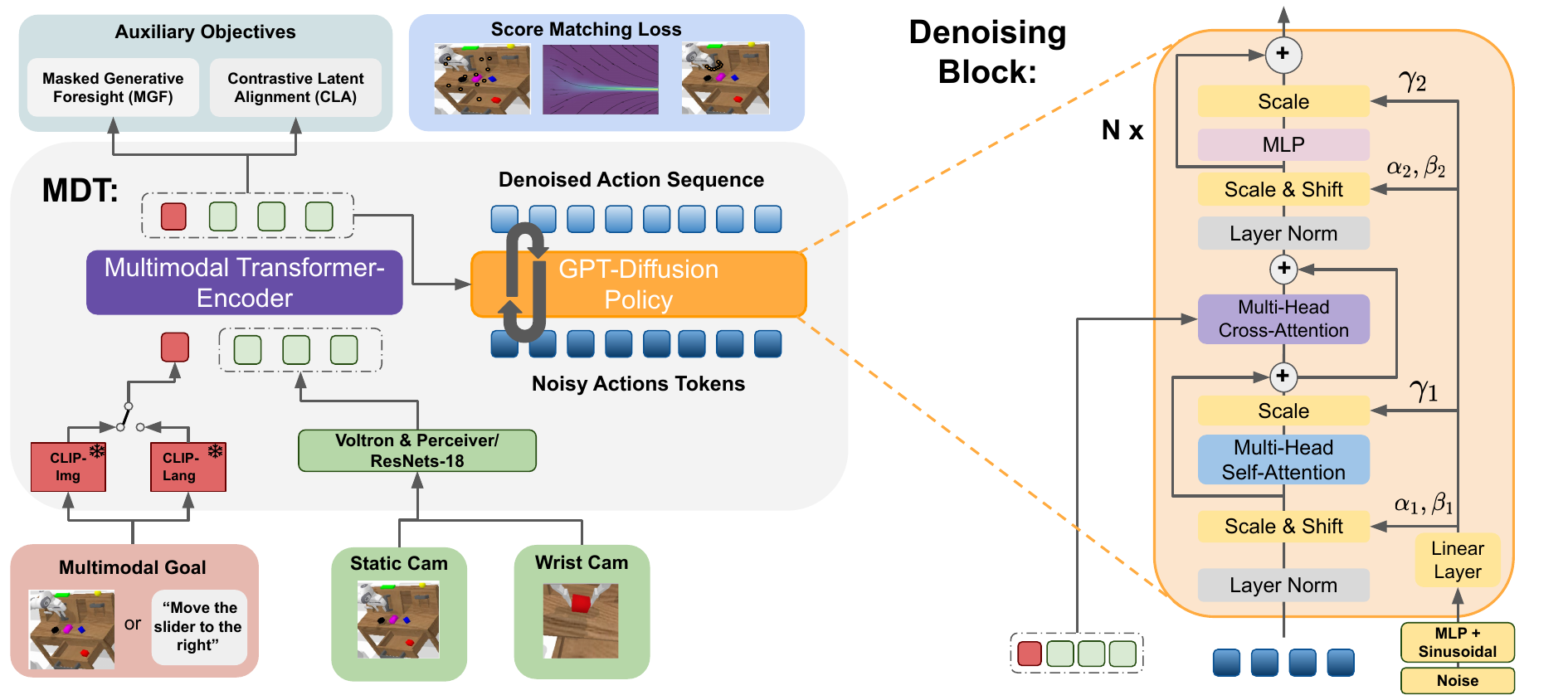}
    \caption{(Left) Overview of the proposed multimodal Transformer-Encoder-Decoder Diffusion Policy used in \gls{mdt}. (Right) Specialized Diffusion Transformer Block for the Denoising of the Action Sequence. 
    \gls{mdt} learns a goal-conditioned latent state representation from multiple image observations and multimodal goals.
    The camera images are processed either via frozen Voltron Encoders with a Perceiver or ResNets.
    The separate GPT denoising module iteratively denoises an action sequence of $10$ steps with a Transformer Decoder with Causal Attention. 
    It consists of several Denoising Blocks, as visualized on the right side. 
    These blocks process noisy action tokens with self-attention and fuse the conditioning information from the latent state representation via cross-attention. 
    \gls{mdt} applies adaLN conditioning~\cite{peebles2023scalable} to condition the blocks on the current noise level. 
    In addition, it aligns the latent representation tokens of the same state with different goal specifications using self-supervised contrastive learning.
    The latent representation tokens are also used as a context input for the masked Image Decoder module to reconstruct masked-out patches from future images. 
    }
    \label{fig: architecture overview}
\end{figure*}

\gls{mdt} sets new standards on CALVIN~\cite{mees2022calvin}, a popular benchmark for language-guided learning from play data comprised of human demonstrations with few language annotations. 
Remarkably, \gls{mdt} requires fewer than $10\%$ of the trainable parameters and no additional pretraining on large-scale datasets to achieve an average $15\%$ absolute performance gain in two CALVIN challenges.
In addition, \gls{mdt} performs exceptionally on the LIBERO benchmark that consists of $5$ task suites featuring $130$ different tasks in several environments. 
To show the efficiency of \gls{mdt}, the tasks are modified such that only $2\%$ of the demonstrations contain language labels.
The results show that \gls{mdt} is even competitive to state-of-the-art methods, that are trained on the fully annotated dataset.
Through a series of experiments and ablations, the efficiency of the method and the strategic design choices are thoroughly evaluated. 
The major contributions of this paper are threefold:
\begin{itemize}
    \item We introduce Multimodal Diffusion Transformer, a novel Transformer-based Diffusion approach. 
    \gls{mdt} excels in learning from multimodal goals and sets a new state-of-the-art performance on the CALVIN Challenge and across all LIBERO task suites.
    \item Two simple yet effective self-supervised losses for visuomotor polices to improve learning from multimodal goals. 
    \gls{mgf} and \gls{cla} improve the performance of multi-task behavior learning from sparsely labeled datasets without additional inference costs.
    \item A comprehensive empirical study covering over $184$ different tasks across several benchmarks, verifying the performance and effectiveness of \gls{mgf} and \gls{mdt}.
\end{itemize}

\section{Related Work}
\paragraph{Language-Conditioned Robot Learning}
Language serves as an intuitive and understandable interface for human-robot interactions, prompting a growing interest in language-guided learning methods within the robotics community.
A growing body of work uses these models as feature generators for vision and language abstractions for downstream policy learning~\cite{shridhar2023perceiver, mees2023grounding, bharadhwaj2023roboagent, mees2022hulc, lynch2023interactive, chen2023playfusion, xian2023unifying} and improved language expression-grounding~\cite{gkanatsios2023energybased, jain2022bottom, rana2023sayplan, xiao2022robotic, moo2023arxiv}. 
Notably, methods like CLIPPort~\cite{shridhar2022cliport} employ frozen CLIP embeddings for language-guided pick and place, while others, such as PaLM-E~\cite{driess2023palm} and RoboFlamingo~\cite{li2023vision}, fine-tune vision-language models for robot control. 
Other methods focus on hierarchical skill learning for language-guided manipulation in \gls{lfp}~\cite{mees2022calvin, lynch2020learning, mees2023grounding, rosete-beas2022latent, zhou2023language, wang2023mimicplay}.
Further, transformer-based methods without hierarchical structures~\cite{zitkovich2023rt, cui2023from, reuss2023goal, wu2023unleashing, shafiullah2022behavior, scheikl2023movement}, focus on next-action prediction based on previous observation tokens.
\gls{mtact}, for instance, utilizes a \gls{vae} transformer encoder-decoder policy, encoding only the current state and a language goal to generate future actions~\cite{bharadhwaj2023roboagent, fu2024mobile, zhao2023learning}.

Furthermore, diffusion-based methods have gained adoption as policy representations that iteratively diffuse actions from Gaussian noise~\cite{ho2020denoising, song2020score}. 
Several diffusion policy approaches focus on generating plans on different abstraction levels for behavior generation.
LAD~\cite{zhang2022lad} trains a diffusion model to diffuse a latent plan sequence in pre-trained latent spaces of HULC~\cite{mees2022hulc} combined with HULC's low-level policy. 
UniPy~\cite{du2023learning} and AVDC~\cite{Ko2023Learning} directly plan in the image space using video diffusion models and execute the plan with another model.
Frameworks related to \gls{mdt} are Distill-Down~\cite{ha2023scaling} and Play-Fusion~\cite{chen2023playfusion}, which also utilize diffusion policies for language-guided policy learning. 
Both methods use variants of the CNN-based diffusion policy~\cite{chi2023diffusionpolicy}.
However, all these methods require fully annotated datasets to learn language-conditioned policies. 
\gls{mdt} effectively learns from multimodal goals, enabling it to leverage partially annotated datasets.

\paragraph{Self-Supervised Learning in Robotics}
An increasing body of work in robotics studies self-supervised representations for robot control.
A key area is learning universal vision representations or world-models, typically trained on large, diverse offline datasets using either masking strategies~\cite{karamcheti2023language, he2022masked, geng2022multimodal, xiao2022masked, seo2023multi, majumdar2023where} or contrastive objectives~\cite{grill2020bootstrap, pari2021surprising, shafiullah2023bringing, zhan2022learning, nair2022r3m, laskin2020curl, becker2023reinforcement, ma2022vip, rana2023contrastive}.
Another body of work explores robust representations for robot policies from multiple sensors, using token masking strategies~\cite{radosavovic2023real} or generative video generation~\cite{wu2023unleashing}.
However, these methods require specific transformer models that rely on a long history of multiple states, which is a limitation for token masking and video generation techniques.
Notably, Crossway-Diffusion~\cite{li2023crossway} proposes a self-supervised loss specifically designed for CNN-based diffusion policies~\cite{chi2023diffusionpolicy} by redesigning the latent space of the U-net diffusion model to reconstruct the current image observation and proprioceptive features for better single task performance.

In order to predict a sequence of future actions efficiently, some recent approaches deploy transformer-based policies that encode only the current state information without any history of prior states~\cite{bharadhwaj2023roboagent, fu2024mobile, zhao2023learning}. 
Traditional token masking strategies~\cite{radosavovic2023real} or video generation objectives~\cite{wu2023unleashing} that rely on token sequences of multiple states for pretraining are incompatible with such single state models, since they rely on a history of prior states.
To bridge this gap, the proposed \gls{mgf} and \gls{cla} objectives enhance the capabilities of these single-state observation policies.
\gls{mgf} and \gls{cla} enable learning of versatile behavior from multimodal goals efficiently and without additional inference costs and can also be used for pretraining on action-free data.

\paragraph{Behavior Generation from Multimodal Goals}
While recent advancements in goal-conditioned robot learning have predominantly focused on language-guided methods, there is a growing interest in developing agents capable of interpreting instructions across different modalities, such as goal images, sketches, and multimodal combinations. 
Mutex~\cite{shah2023mutex} presents an imitation learning policy that understands goals in natural speech, text, videos, and goal images.
Mutex further uses cross-modality pretraining to enhance the model's understanding of the different goal modalities.
Steve-1~\cite{lifshitz2023steve} is a Minecraft agent that uses a \gls{vae} encoder to translate language descriptions into the latent space of video demonstrations of the task, enabling it to follow instructions from both videos or text descriptions. 
Other research efforts are exploring novel conditioning methods. 
Examples include using the cosine distance between the current state and a goal description from fine-tuned CLIP models~\cite{seo2023multi} or employing multimodal prompts~\cite{jiang2023vima} that combine text with image descriptions.
Rt-Sktech and Rt-Trajectory present two new conditioning methods leveraging goal sketches of the desired scene~\cite{zitkovich2023rt} and sketched trajectories of the desired motion~\cite{gu2023rttrajectory}, respectively.
While \gls{mdt} primarily addresses the two most prevalent goal modalities, namely text and images, our framework is in theory versatile enough to incorporate other modalities like sketches.

\section{Method}

\gls{mdt} is a diffusion-based transformer encoder-decoder architecture that simultaneously leverages two self-supervised auxiliary objectives.
Namely Contrastive Latent Alignment and Masked Generative Foresight.
First, the problem definition is provided.
Next, the continuous-time diffusion formulation, essential for understanding action sequence learning from play, of \gls{mdt} is discussed. 
Followed by an overview of the proposed transformer architecture of \gls{mdt}. 
Afterward, the two novel self-supervised losses are introduced.

\subsection{Problem Formulation}
The goal-conditioned policy $\pi_{\theta}(\act | \state, \goal)$ predicts a sequence of actions $\act = (\act*,\ldots,\act*[i+k-1])$ of length $k$, conditioned on both the current state embedding $\state$ and a latent goal $\goal$. 
The latent goal $\goal \in \{\goalI,\mathbf{l}\}$ encapsulates either a goal-image $\mathbf{o}$ or an encoded free-form language instruction $\goalL$. 
\gls{mdt} learns such policies from a set of task-agnostic play trajectories $\Tset$.
Each individual trajectory $\tau \in \Tset$ represents a series of tuples $\tau = \left((\state[1], \act*[1]),\ldots,(\state[T_n], \act*[T_n])\right)$, with observation $\state$, action $\act*$.
The final play dataset is defined as $\Dset = \left\{(\state, \act)\middle| \act = (\act*,\ldots,\act*[i+k-1]),(\state, \act*)\in\tau, \tau \in \Tset\right\}$.
During training, a set of goals is created for each datapoint $\goalSet_{\state,\act}=\{\goalI_i,\goalL_i\}$, where $\goalL_i$ is the language annotation for the state $\state$ if it exists in the dataset. 
The goal image $\goalI_i = \state[i+j]$ is a future state where the offset $j$ is sampled from the geometric distribution with bounds $j\in[20,50]$ and probability of $0.1$.
\gls{mdt} maximizes the log-likelihood across the play dataset,
\begin{align}
\mathcal{L}_{\text{play}} = \mathbb{E}\left[ \sum_{(\state, \act) \in \Dset}\sum_{\goal\in\mathcal{G}_{\state,\act}} \log \p{\pi_\theta}{\act | \state, \goal} \right].
\end{align}
Human behavior is diverse and there commonly exist multiple trajectories converging towards an identical goal.
The policy must be able to encode such versatile behavior~\cite{blessing2023information} to learn effectively from play.

\subsection{Score-based Diffusion Policy}
In this section, the language-guided Diffusion Policy for learning long-horizon manipulation from play with limited language annotations is introduced.
Diffusion models are generative models that learn to generate new data from random Gaussian noise through an iterative denoising process.
The models are trained to subtract artificially added noise with various noise levels.
Both the procedures of adding and subtracting noise can be described as continuous time processes stochastic-differential equations (SDEs)~\cite{song2020score}. 
\gls{mdt} leverages a continuous-time SDE formulation~\cite{karras2022elucidating}
\begin{equation}
\label{eq: conditional Karras Song SDE}
\diff \act  =  \big( \beta_t \sigma_t - \dot{\sigma}_t  \big) \sigma_t \nabla_a \log p_t(\act | \state, \goal) dt + \sqrt{2 \beta_t} \sigma_t d\omega_t,
\end{equation}
that is commonly used in image generation~\cite{karras2022elucidating, song2023consistency}.
The score-function $\nabla_{\act} \log p_t(\act | \state, \goal)$ is parameterized by the continuous diffusion variable $t \in [0, T]$, with constant horizon $T > 0$.
This formulation reduces the stochasticity to the Wiener process $\omega_t$, which can be interpreted as infinitesimal Gaussian noise that is added to the action sample. 
The noise scheduler $\sigma_t$ defines the rate of added Gaussian noise depending on the current time $t$ of the diffusion process. 
Following best practices~\cite{karras2022elucidating, reuss2023goal, song2023consistency}, \gls{mdt} uses $\sigma_t=t$ for the policy. 
The range of noise perturbations is set to  $\sigma_t \in [0.001, 80]$ and the action range is rescaled to $[-1, 1]$.
The function $\beta_t$ describes the replacement of existing noise through injected new noise \cite{karras2022elucidating}.
This SDE is notable for having an associated ordinary differential equation, the Probability Flow ODE~\cite{song2020score}. 
When action chunks of this ODE are sampled at time $t$ of the diffusion process, they align with the distribution $p_t(\act| \state, \goal)$,
\begin{equation}
\label{eq: karras ODE}
\diff \act = - t \nabla_{\act} \log p_t(\act| \state, \goal)\diff t.
\end{equation}

The diffusion model learns to approximate the score function $\nabla_{\act} \log p_t(\act |\state, \goal)$ via Score matching (SM)~\cite{6795935}
\begin{equation}
\label{eq: SM loss}
    \mathcal{L}_{\text{SM}} = \mathbb{E}_{\mathbf{\sigma}, \act, \boldsymbol{\epsilon}} \big[ \alpha (\sigma_t) \newline  \| D_{\theta}(\act + \boldsymbol{\epsilon}, \state, \goal, \sigma_t)  - \act  \|_2^2 \big],
\end{equation}
where $D_{\theta}(\act + \boldsymbol{\epsilon}, \state, \goal, \sigma_t)$ is the trainable neural network.
During training, noise levels from a noise distribution $p_{\text{noise}}$ are sampled randomly and added to the action sequence and the model predicts the denoised action sequence. 
To generate actions during a rollout, the learned score model is inserted into the reverse SDE and the model iteratively denoises the next sequence of actions.
By setting $\beta_t = 0$, the model recovers the deterministic inverse process that allows for fast sampling in a few denoising steps without injecting additional noise into the inverse process~\cite{song2020score}.
Detailed training and inference description can be found in~\autoref{sec: app score training and inference} of the Appendix.
For the experiments, \gls{mdt} uses the DDIM sampler~\cite{song2020score} to diffuse an action sequence in $10$ steps.

\subsection{Model Architecture}
\begin{figure}
    \centering
    \includegraphics[width=0.48\textwidth]{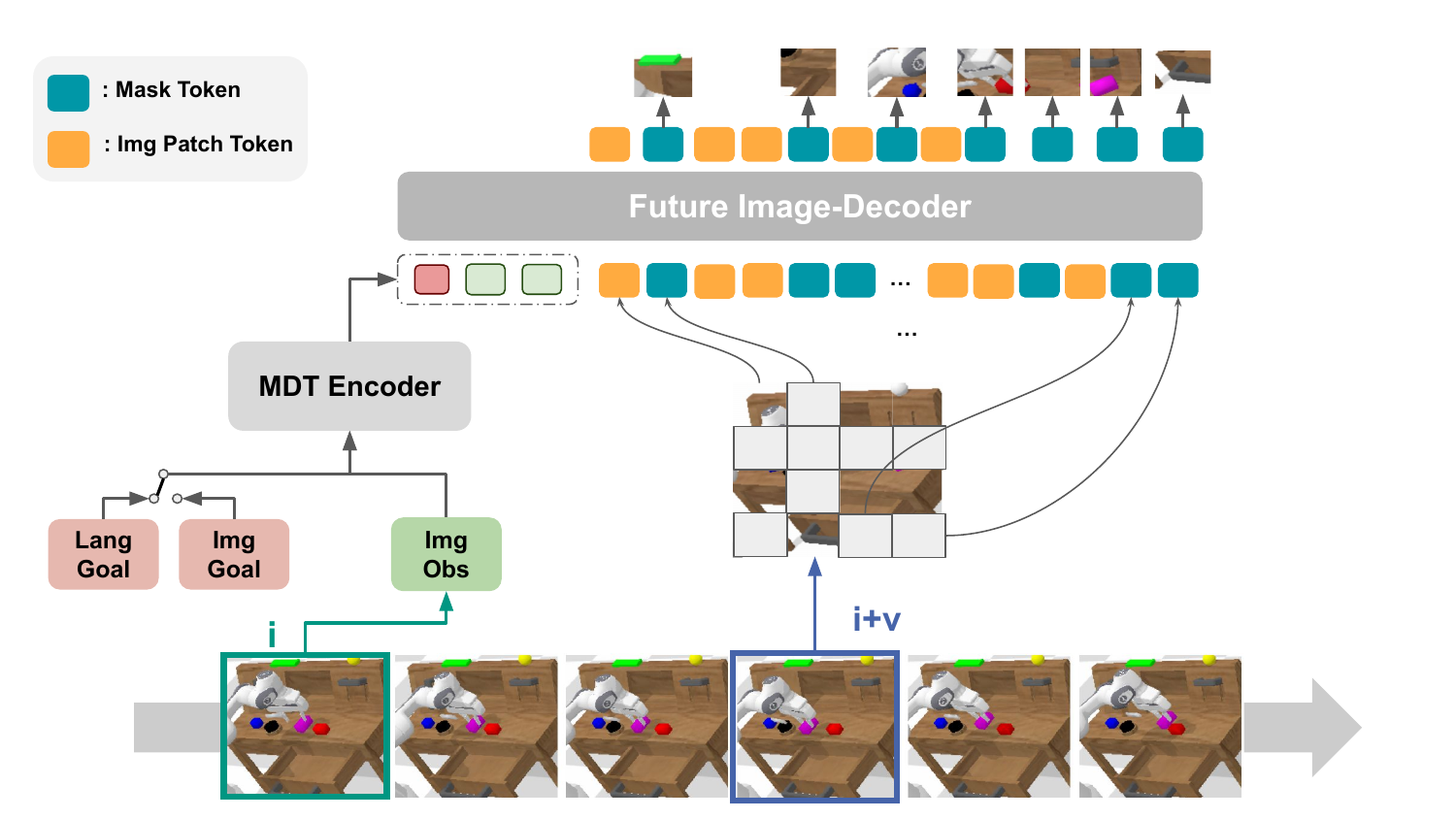}
    \caption{The Masked Generative Foresight Auxiliary Task enhances the \gls{mdt} model. 
    It starts by encoding the current observation and goal using the \gls{mdt} Encoder. 
    The resulting latent state representations then serve as conditional inputs for the Future Image-Decoder. 
    This decoder receives encoded patches of future camera images along with mask tokens. 
    Its task is to reconstruct the occluded patches in future frames.}
    \label{fig:enter-label}
\end{figure}
\gls{mdt} uses a multimodal transformer encoder-decoder architecture to approximate the conditional score function of the action sequence.
The encoder first processes the tokens from the current image observations and desired multimodal goals, converting these inputs into a series of latent representation tokens.
The decoder functions as a diffuser that denoises a sequence of future actions.
\autoref{fig: architecture overview} illustrates the architecture.

First, \gls{mdt} encodes image observations of the current state from multiple views with image encodings. 
This work introduces two encoder versions of \gls{mdt}: \textit{\gls{mdt}-V}, a variant with the frozen Voltron embeddings and \textit{\gls{mdt}}, the default model with ResNets.
The \gls{mdt}-V encoder leverages a Perceiver-Resampler to improve computational efficiency~\cite{alayrac2022flamingo}.
Each image is embedded into $196$ latent tokens by Voltron.
The Perceiver module uses multiple transformer blocks with cross attention to compress these Voltron tokens into a total of $3$ latent tokens.
This procedure results in a highly efficient feature extractor that capitalizes on pretrained Voltron embeddings.
\begin{figure*}
    \centering
    \includegraphics[width=\linewidth]{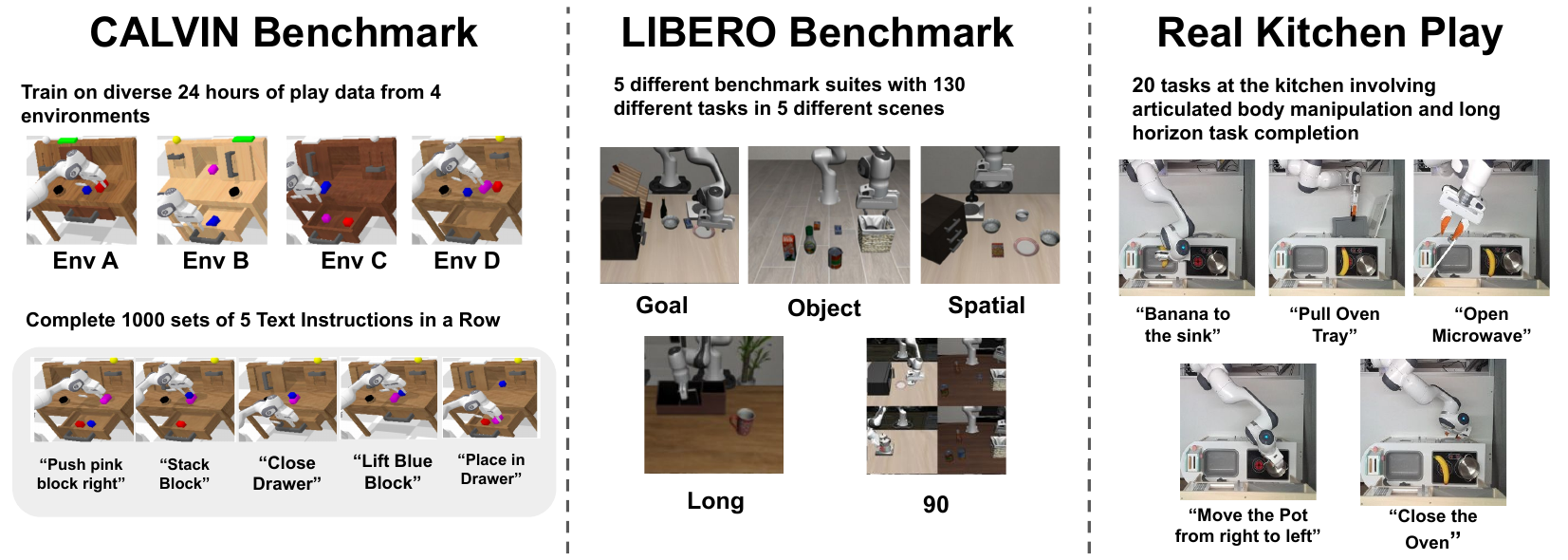}
    \caption{Overview of the different environments used to test \gls{mdt}: (Left) CALVIN Benchmark consisting of four environments each with unique positions and textures for slider, drawer, LED, and lightbulb.  
    (Middle) Overview of the different tasks and scene diversity in the LIBERO benchmark, which is divided into $5$ different task suites.
    (Right) Example tasks from the real robot experiments at a toy kitchen, where models are tested after training on partially labeled play data.}
    \label{fig: sim_overview}
\end{figure*}
The \gls{mdt} encoder uses a trainable ResNet-18 with spatial softmax pooling and group norm~\cite{chi2023diffusionpolicy} for each camera view.
Each ResNet returns a single observation token for every image. 
Both \gls{mdt} encoder versions embed goal images and language annotations via frozen CLIP models~\cite{radford2021learning} per goal-modality into a single token.
After the computation of the embeddings, both MDT encoders apply the same architecture comprised of several self-attention transformer layers, resulting in a set of informative latent representation tokens.

The \gls{mdt} diffusion decoder denoises the action sequence with causal masking. 
Cross-attention in every decoder layer fuses the conditioning information from the encoder into the denoising process.
The current noise level $\sigma_t$ is embedded using a Sinusoidal Embedding with an additional MLP into a latent noise token. 
\gls{mdt} conditions the denoising process to the current noise level via AdaLN-conditioning on the Transformer Decoder blocks~\cite{peebles2023scalable}.
The right part of~\autoref{fig: architecture overview} illustrates this process, including all internal update steps.
The proposed framework separates representation learning from denoising, resulting in a more computationally efficient model since the model only needs to encode the latent representation tokens once.
Further, the experiments demonstrate that the proposed denoising model achieves higher performance than prior Diffusion-Transformer architectures~\cite{chi2023diffusionpolicy}.
\gls{mdt} uses the same set of hyperparameters across all experiments.

\subsection{Masked Generative Foresight}
A key insight of this work is that policies require an informative latent space to understand how desired goals will change the robot's behavior in the near future.
Consequently, policies that are able to follow multimodal goals have to map different goal modalities to the same desired behaviors. 
Whether a goal is defined through language or represented as an image, the intermediate changes in the environment are identical across these goal modalities. 
The proposed \textit{Masked Generative Foresight}, an additional self-supervised auxiliary objective, builds upon this insight. 
Given the latent embedding of the \gls{mdt}(-V) encoder for state $\state$ and goal $\goal$, \gls{mgf} trains a \gls{vit} to reconstruct a sequence of 2D image patches $\left(\patch_1,\ldots,\patch_{\Npatch}\right) =\mathtt{patch}(\state[i+v])$ of the future state $\state[i+v]$, with $v=3$ being the foresight distance used across all experiments in this work. 
A random subset of $\maskpercent\%$ of these patches is replaced by a mask token. 
Even though the \gls{vit} now receives both masked and non-masked patches only the reconstruction of the masked patches contributes to the loss
\begin{equation}
\label{eq: mgf loss}
    \mathcal{L}_{\mathrm{MGF}}\left(\state\right) = \frac{1}{\Npatch}\sum_{\mathclap{\patch\in\mathtt{patch}(\state[i+v])}}\boldsymbol{\mathrm{1}}_\mathrm{m}(\patch)\left(\patch-\hat{\patch}\right)^2,
\end{equation}
where the indicator function $\boldsymbol{\mathrm{1}}_\mathrm{mk}(\patch)$ is $1$ if $\patch$ is masked and $0$ otherwise.
Detailed hyperparameters for the model are summarized in \autoref{tab: MGF model hyperparameters} of the Appendix.

\gls{mgf} differs from existing approaches~\cite{li2023crossway, wu2023unleashing}, which require full reconstruction of images or videos.
While various other masking methods exist~\cite{karamcheti2023language}, all of them aim to learn robust representations of the current state, while \gls{mgf} reconstructs future states to include foresight into the latent embedding.
\gls{mgf} is conceptually simple and can be universally applied to all transformer policies.
Section \ref{subsec: EvalAuxLosses} shows that the advantages of \gls{mgf} are not specific to \gls{mdt} but also increases the performance of \gls{mtact}~\cite{bharadhwaj2023roboagent}.

\subsection{Aligning Latent Goal-Conditioned Representations}

To effectively learn policies from multimodal goal specifications, \gls{mdt} must align visual goals with their language counterparts. A common approach to retrieve aligned embeddings between image and language inputs is the pre-trained CLIP model, which has been trained on paired image and text samples from a substantial internet dataset~\cite{radford2021learning}.
However, CLIP exhibits a tendency towards static images and struggles to interpret spatial relationships and dynamics~\cite{xiao2022robotic, myers2023goal, mees2022hulc}.
These limitations, lead to an insufficient alignment in \gls{mtil} since goal specifications in robotics are inherently linked to the dynamics between the current state $\state$ and the desired goal $\goal$. 
Instead of naively fine-tuning the large 300-million-parameter CLIP model, \gls{mdt} introduces an auxiliary objective that aligns the \gls{mdt}(-V) state embeddings conidtioned on different goal modalities. 
These embeddings include the goal as well as the current state information, allowing the \gls{cla} objective to consider the task dynamics.

Since \gls{cla} requires a single vector for each goal modality, the various \gls{mdt}-V latent tokens are reduced via Multi-head Attention Pooling~\cite{karamcheti2023language} and subsequently normalized. 
\gls{mdt} uses the embedding of the static image as a representative token to compute the contrastive loss.
Hence, every training sample $(\state,\act)$ that is paired with a multimodal goals specification $\goalSet_{\state,\act}=\{\goalI_i,\goalL_i\}$ is reduced to the vectors $\clavar_i^\goalI$ and $\clavar_i^\goalL$ for images and language goals respectively.
\gls{cla} computes the InfoNCE loss using the cosine similarity $\cossim\left(\clavar_i^{\goalI},\clavar_i^{\goalL}\right)$ between the image-goal conditioned state embedding $\clavar_i^{\goalI}$ and the language-goal conditioned state embedding $\clavar_i^{\goalL}$
\begin{align}
\label{eq: clip loss}
\mathcal{L}_{\text{CLA}} = &-\frac{1}{2B} \sum_{i=1}^B \left( \log \left( \frac{\exp\left(\frac{\cossim\left(\clavar_i^\goalI,\clavar_i^\goalL\right)}{\clatemp}\right)}{\sum_{j=1}^B \exp\left(\frac{\cossim\left(\clavar_i^\goalI,\clavar_j^\goalL\right)}{\clatemp}\right)} \right) \right. \nonumber\\
&\left. + \log \left( \frac{\exp\left(\frac{\cossim\left(\clavar_i^\goalI,\clavar_i^\goalL\right)}{\clatemp}\right)}{\sum_{j=1}^B \exp\left(\frac{\cossim\left(\clavar_j^\goalI,\clavar_i^\goalL\right)}{\clatemp}\right)} \right) \right),
\end{align}
with temperature parameter $\clatemp$ and batch size $B$.
The full \gls{mdt} loss combines the Score Matching loss, from Eq. \eqref{eq: SM loss}, the \gls{mgf} loss from Eq. \eqref{eq: mgf loss} and the \gls{cla} loss from Eq. \eqref{eq: clip loss}
\begin{equation}
    \mathcal{L}_{\text{MDT}} = \mathcal{L}_{\text{SM}} +  \alpha \mathcal{L}_{\text{MGF}} + \beta \mathcal{L}_{\text{CLIP}},
\end{equation}
where $\alpha = 0.1$ and $\beta = 0.1$ in most experiment settings.

\section{Evaluation}

In this section, we examine the performance of MDT on CALVIN~\cite{mees2022calvin} and LIBERO~\cite{liu2023libero}, two established benchmarks for Language-conditioned Imitation Learning. MDT is tested against several state-of-the-art methods on both benchmarks. 
In addition, we evaluate MDT in a real world play setting. 
The experiments aim to answer the following questions:
\begin{itemize}
    \item (I) Is \gls{mdt} able to learn long-horizon manipulation from play data with few language annotations?
    \item (IIa) Do \gls{mgf} and \gls{cla} enhance the performance of \gls{mdt}? 
    \item (IIb) Does \gls{mgf} improve the performance of other transformer policies?
    \item (III) Can \gls{mdt} learn language-guided manipulation from partially labeled data in a real-world setting?
\end{itemize}

\subsection{Evaluation on CALVIN}

\begin{table}
\centering
\scalebox{0.75}{
\begin{tabular}{ll|ccccccc}
\toprule
 \multirow{2}{*}{Train} &  \multirow{2}{*}{Method} & \multicolumn{6}{c}{No. Instructions in a Row (1000 chains)}  \\
  \cmidrule(lr){3-8}
 &  & 1 & 2 & 3 & 4 & 5 & \textbf{Avg. Len.} \\ \midrule

 \multirow{6}{*}{D} &  HULC & 82.5\% & 66.8\% & 52.0\% & 39.3\% & 27.5\% & 2.68$\pm$(0.11)  \\
& LAD & 88.7\% & 69.9\% & 54.5\% & 42.7\% & 32.2\% & 2.88$\pm$(0.19) \\
& Distill-D & 86.7\% & 71.5\% & 57.0\% & 45.9\% & 35.6\% & 2.97$\pm$(0.04) \\
& MT-ACT & 88.4\% & 72.2\% & 57.2\% & 44.9\% & 35.3\% & 2.98$\pm$(0.05) \\
& \textbf{MDT (ours)} & \textbf{93.3\%} & \textbf{82.4\%} & \textbf{71.5\%} & \textbf{60.9\%} & \textbf{51.1\%} & \textbf{3.59$\pm$(0.07)}  \\
& \textbf{MDT-V (ours)} & \textbf{93.7\%} & \textbf{84.5\%} & \textbf{74.1\%} & \textbf{64.4\%} & \textbf{55.6\%} & \textbf{3.72$\pm$(0.05)}  \\
\midrule
 \multirow{6}{*}{ABCD} 
& HULC & 88.9\% & 73.3\% & 58.7\% & 47.5\% & 38.3\% & 3.06$\pm$(0.07) \\
& Distill-D & 86.3\% & 72.7\% & 60.1\% & 51.2\% & 41.7\% & 3.16$\pm$(0.06) \\
& MT-ACT & 87.1\% & 69.8\% & 53.4\% & 40.0\% & 29.3\% & 2.80$\pm$(0.03) \\
& RoboFlamingo & 96.4\% & 89.6\% & 82.4\% & 74.0\% & 66.0\% & 4.09$\pm$(0.00) \\
& \textbf{MDT (ours)} & \textbf{97.8\%} & \textbf{93.8\%} & \textbf{88.8\%} & \textbf{83.1\%} & \textbf{77.0\%} & \textbf{4.41$\pm$(0.03)} \\
& \textbf{MDT-V (ours)} & \textbf{98.6\%} & \textbf{95.8\%} & \textbf{91.6\%} & \textbf{86.2\%} & \textbf{80.1\%} & \textbf{4.52$\pm$(0.02)} \\
\bottomrule
\end{tabular}}

\caption{
Performance comparison of various policies learned end-to-end on the CALVIN ABCD$\rightarrow
$D and D$\rightarrow$D challenge within the CALVIN benchmark. 
The table shows the average rollout length to solve $5$ instructions in a row (Avg. Len.) of $1000$ chains. 
The proposed methods MDT and MDT-V significantly outperform all reported baselines averaged over 3 seeds on both datasets and sets a sota performance.
}
\label{tab: CALVIN Long horizon}
\end{table}
\begin{table*}[h!]
\centering
\scalebox{0.95}{
\begin{tabular}{l|c|cc|cccccc}
\toprule
Method & Lang. Annotation & $\mathcal{L}_{\text{CLA}}$ & $\mathcal{L}_{\text{MGF}}$ & Spatial & Object & Goal  & Long & 90 & Average  \\ 
\midrule
Transformer-BC~\cite{liu2023libero} & 100 $\%$ & $\times$ &  $\times$&  71.8 $\pm$ 3.7 & 71.00 $\pm$ 7.9 & \textbf{76.3} $\pm$ \textbf{1.3} &  $24.2 \pm 2.6$ & - & -  \\
\midrule
 Distill-D~\cite{ha2023scaling} & $2\%$ & $\times$ & $\times$ & $46.8\pm2.8$   & $72.0 \pm 6.5$   & $63.8 \pm 2.5$ & $47.3 \pm 4.1$ & $49.9 \pm 1.0$  & $56.0 \pm 3.4$ \\
\midrule
\multirow{4}{*}{MDT}  & $2\%$ & $\times$ & $\times$ & $66.0 \pm 1.9$   & $85.2 \pm 2.3$   & $67.8 \pm 4.6$ & $65.0 \pm 2.0$ & $58.7 \pm 0.8$  & $68.5 \pm 9.92$ \\
& $2\%$ & $\checkmark$ & $\times$ & $74.3 \pm 0.8$   & $87.5 \pm 2.7$   & $71.5 \pm 3.5$ &  \textbf{65.3} $\pm$ \textbf{2.1} & $66.9 \pm 1.7$  &  $73.1 \pm 8.81$ \\
& $2\%$ & $\times$ & $\checkmark$ &  $67.5 \pm 2.1$   & $87.5 \pm 2.6$   & $69.3 \pm 2.5$ &  $63.0 \pm 1.7$ & $62.6 \pm 1.0$  & $70.0 \pm 10.2$ \\
&  $2\%$ & $\checkmark$ & $\checkmark$ & \textbf{78.5} $\pm$ \textbf{1.5}  & \textbf{87.5} $\pm$ \textbf{0.9}   & 73.5 $\pm$ 2.0 & $64.8 \pm 0.3$ & \textbf{67.2} $\pm$ \textbf{1.1} & \textbf{74.3} $\pm$ \textbf{9.13} \\ 
\bottomrule
\end{tabular}
}
\caption{Overview of the performance of MDT and baselines with and without our proposed Self-Supervised Losses on several LIBERO Task suites. 
All results show the average performance of all tasks averaged over 20 rollouts each and with 3 seeds. 
MDT does outperform the Transformer-BC baseline in several settings with only $2\%$ of language annotations.}
\label{tab: libero ablations}
\end{table*}
The CALVIN challenge~\cite{mees2022calvin} 
consists of four similar but different environments A, B, C, D. 
The four setups vary in desk shades and the layout of items as visualized in \autoref{fig: sim_overview}.
The main experiments for this benchmark are conducted on the full dataset ABCD$\rightarrow$D, where the policies are trained on ABCD and evaluated on D. 
This setting contains $24$ hours of uncurated teleoperated play data with multiple sensor modalities and $34$ different tasks for the model to learn. 
Further, only $1\%$ of data is annotated with language descriptions.
All methods are evaluated on the long-horizon benchmark, which consists of $1000$ unique sequences of instruction chains, described in natural language. 
Every sequence requires the robot to continuously solve $5$ tasks in a row.
During the rollouts, the agent gets a reward of $1$ for completing the instruction with a maximum of $5$ for every rollout.
We additionally perform experiments on the small benchmark D$\rightarrow$D consisting of only 6 hours of play data to study the data efficiency of our proposed method.

\textbf{Baselines.}
We compare our proposed policy against the following state-of-the-art language-conditioned multi-task policies on CALVIN:
\begin{itemize}
    \item \textbf{HULC:} A hierarchical play policy, that uses discrete \gls{vae} skill space with an improved low-level action policy and a transformer plan encoder to learn latent skills \cite{mees2022hulc}.
    \item \textbf{LAD:} A hierarchical diffusion policy, that extends the HULC policy by substituting the high-level planner with a U-Net Diffusion model~\cite{zhang2022lad} to diffuse plans. 
    \item \textbf{Distill-D:} A language-guided Diffusion policy from~\cite{ha2023scaling}, that extends the initial U-Net diffusion policy~\cite{chi2023diffusionpolicy} with additional Clip Encoder for language-goals. 
    We use our continuous time diffusion variant instead of the discrete one for direct comparison and extend it with the same CLIP vision encoder to guarantee a fair comparison. 
    \item \textbf{MT-ACT:} A multitask transformer policy~\cite{bharadhwaj2023roboagent, zhao2023learning}, that uses a \gls{vae} encoder for action sequences and also predicts action chunks instead of single actions with a transformer encoder-decoder architecture. 
    \item \textbf{RoboFlamingo:} A finetuned Vision-Language Foundation model \cite{li2023vision} containing $3$ billion parameters, that has an additional recurrent policy head for action prediction.
    The model was pretrained on a large internet-scale set of image and text data and then finetuned for CALVIN. 
\end{itemize}
We adopt the recommended hyperparameters for all baselines to guarantee a fair comparison and give an overview of our chosen hyperparameters for self-implemented baselines in Appendix~\ref{sec: baseline Implementations}.
Further, we directly compare the self-reported results of HULC, LAD, and RoboFlamingo on CALVIN \cite{zhang2022lad, mees2022hulc, li2023vision, zhang2022lad}. 
All models use the same language and image goal models to ensure fair comparisons.
Since RoboFlamingo only published the best seed of each model, we can not include standard deviations in their results.

\textbf{Results.}
The results of all our experiments on CALVIN are summarized in Table~\ref{tab: CALVIN Long horizon}. 
We assess the performance of \gls{mdt} and \gls{mdt}-V on ABCD$\rightarrow$D and on the small subset D$\rightarrow$D.
The results are shown in \autoref{tab: CALVIN Long horizon}. 
\gls{mdt}-V sets a new record in the CALVIN challenge, extending the average rollout length to $4.52$ which is a $10\%$ absolute improvement over RoboFlamingo. 
\gls{mdt} also surpasses all other tested methods.
Notably, \gls{mdt} achieves this while having less than $10\%$ of trainable parameters and not requiring pretraining on large-scale datasets. 
In the scaled-down CALVIN D$\rightarrow$D benchmark, \gls{mdt}-V establishes a new standard, outperforming recent methods like LAD~\cite{zhang2022lad} and boosting the average rollout length by $20\%$ over the second best baseline.
While RoboFlamingo demonstrates commendable performance on the complete ABCD dataset, it relies on substantial training data and remains untested on the D$\rightarrow$D subset.
In contrast, \gls{mdt} excels in both scenarios with remarkable efficiency.

\subsection{Evaluation on LIBERO}

We further evaluate various models on LIBERO~\cite{liu2023libero}, a robot learning benchmark consisting of over 130 language-conditioned manipulation tasks divided into $5$ different task suites: LIBERO-Spatial, LIBERO-Goal, LIBERO-Object, LIBERO-90, and LIBERO-Long.
Each task suite except for LIBERO-90 consists of $10$ different tasks with $50$ demonstrations each.
To evaluate the ability of \gls{mdt} to effectively learn from partially labeled data, we only label a fraction of $2\%$ with the associated task description.
Each task suite focuses on different challenges of imitation learning: LIBERO-Goal tests on tasks with similar object categories but different goals. LIBERO-Spatial requires policies to adapt to changing spatial arrangements of the same objects. In contrast, LIBERO-Object maintains he layout while changing the objects.
LIBERO-90 is the only suite that consists of $90$ different tasks in several diverse environments and tasks with various spatial layouts. 
During evaluation, we test all models on all tasks with $20$ rollouts each and average the results over $3$ seeds.
During the experiments, we restrict all policies to only use a static camera and a wrist-mounted one.
Further details for all task suites are provided in \autoref{sec: libero} of the Appendix.

\textbf{Baselines.}
For our experiments in LIBERO, we report the performance of MDT, Distill-D and the best transformer baseline policy from the original benchmark, which was trained with full language annotations~\cite{liu2023libero}.

\textbf{Results.}
In the LIBERO task suites, summarized in Table~\ref{tab: libero ablations}, \gls{mdt} proves to be effective with sparsely labeled data, outperforming the Oracle-BC baseline, which relies on fully labeled demonstrations.
\gls{mdt} not only outperforms the fully language-labeled Transformer Baseline in three out of four challenges but also significantly surpasses the U-Net-based Distill-D policy in all tests by a wide margin, even without auxiliary objectives. 
The proposed auxiliary objectives further improve the average performance of \gls{mdt} by $8.5\%$ averaged over all $5$ task suites.
These outcomes highlight the robustness of our architecture and affirmatively answer Question (I) regarding its efficiency.

\subsection{Real Robot Experiments}
We investigate research question (III) by assessing the ability of \gls{mdt} to learn language-guided manipulations from partially labeled data in a real-world setting.

\textbf{Robot Setup.}
\gls{mdt} is evaluated on a real-world play kitchen setup with a 7 DoF Franka Emika Panda Robot.
The toy kitchen has an oven, a microwave, a cooler and a sink.
In addition, we positioned a toaster, a pot and a banana in the environment for the robot to interact with.
A detailed overview of our setup is given in \autoref{fig:pdf-comparison} in the Appendix.
The setup incorporates two static RGB cameras: one positioned above the kitchen for a bird's-eye view, and another placed on the right side of the robot.
The action space is the normalized joint space, $[-1, 1]$, of the robot and the binary gripper control.

\textbf{Play Dataset.}
The real-world play dataset encompasses around $4.5$ hours of interactive play data with $20$ different tasks for the policies to learn.
Long-horizon demonstrations consisting of several tasks have been collected by volunteers via teleoperation.
The volunteers were not given any instructions on how many tasks their demonstration should contain, which tasks they should perform, in which order tasks should be performed or which object they should interact with. 
The resulting demonstrations vary greatly in their duration and hence the number of contained tasks. The demonstrations last from around 30 seconds to more than 450 seconds, and contain between 5 and 20 tasks.
The dataset is partially labeled by randomly identifying some tasks in the demonstrations and annotating the respective interval, yielding a total of $360$ labels or approximately $20\%$ of the dataset.
Stationary states at the beginning and end of each demonstration were trimmed and the camera view was cropped to exclude the teleoperator from the images. Other than these adjustments no additional preprocessing was performed on the demonstrations.
Hence, the methods have to learn multimodal goal-conditioned policies from partially labeled, unsegmented, long-horizon play data. Training a single agent to perform $20$ different skills from such a dataset is a very hard challenge for the tested approaches.
More detailed descriptions of all $20$ tasks with additional visualizations are provided in \autoref{fig:real-robot-tasks} in the Appendix.

\textbf{Single Task Evaluation.}
First, we test several policies to complete various single tasks from the play dataset. 
We test all policies on a single task setting with $5$ rollouts per task. 
During each rollout the robot starts from a central, randomized starting position, which was not used in training.
Human observers decide if a task was solved successfully for each rollout.
We compare \gls{mdt} with our proposed auxiliary objectives against \gls{mdt} trained without them and against MT-ACT. The goals were given as language instructions. The results are averaged over $5$ rollouts for each of the $20$ tasks. The results are summarized in Table \ref{tab: real robot avg summary}. The poor performance of $0.25$ success rate for a strong state-of-the-art baseline such as MT-ACT highlights how challenging this setting is. In comparison, MDT achieves a respectable $0.51$ success rate without and $0.58$ with the auxiliary objectives. This improvement is consistent with the evaluations on the LIBERO benchmark.

\begin{table}[]
    \centering
    \begin{tabular}{l|c}
        Model & Avg. Single Task \\
        \midrule
        MT-ACT & 0.25 $\pm$ 0.43 \\
        MDT & 0.51 $\pm$ 0.50 \\
        MDT + $\mathcal{L}_{\text{MGF}}$ + $\mathcal{L}_{\text{CLA}}$ & \textbf{0.58} $\pm$ \textbf{0.49} \\
    \bottomrule
    \end{tabular}
    \caption{The single task performance on the real world dataset with language-conditioned goals. The average is computed over $5$ rollouts for each of the $20$ tasks. The poor performance of MT-ACT showcases the difficulty of this experiment. \gls{mdt} performs significantly better with an additional boost through the auxiliary objectives. and \gls{mdt} $+$ auxiliary losses.}
    \label{tab: real robot avg summary}
\end{table}

\textbf{Long-Horizon Multi-Task Evaluation.}
Finally, we test the approaches to complete several instructions in a single sequence. 
This requires policies to chain different behaviors together.
An example instruction chain is: "Push toaster", "Pickup toast and put it to sink", "Move banana from right stove to sink", "Move pot from left to right stove", "Open oven", "Open microwave".
During a rollout, the model only gets the next goal, if it has completed the prior task successfully.
We test all policies using $4$ different instruction chains.
These instruction chains are detailed in the Appendix.
We evaluate language and image goal-conditioning and report the average rollout length over all chains averaged over $4$ rollouts each. The results are summarized in Table \ref{tab: real robot experiments}. Given the increased complexity of this experiment over the single task evaluation the modest performance of all three methods is not surprising, further highlighting the difficulty of this setting. Especially when considering, that the entire dataset only encompasses $4.5h$ of unsegmented play-data.
MT-ACT again performs significantly worse than \gls{mdt}, with an average rollout length of $0.81$ vs $1.38$ and $0.13$ vs $0.56$ for the language and image goals respectively. Moreover, the experiment shows an even stronger contribution of the proposed auxiliary objectives to the overall performance compared to the single task evaluation.
Further details are discussed in the Appendix.
\begin{figure}
    \centering
    \includegraphics[width=\linewidth]{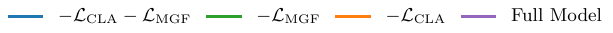}
    \includegraphics[width=\linewidth]{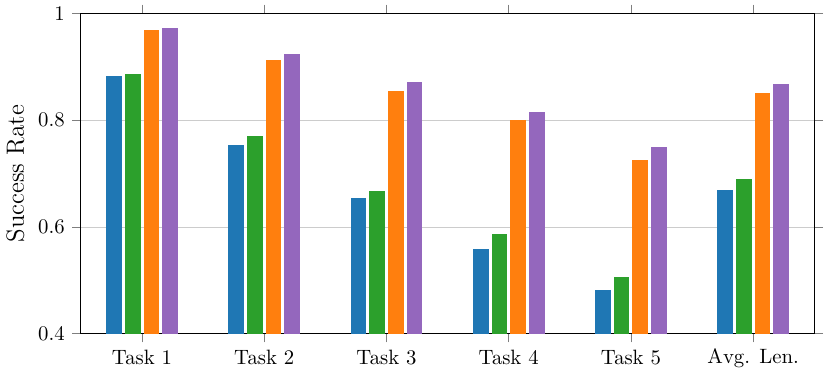}
    \caption{Study on the performance of our proposed Masked Generative Foresight Loss and the Contrastive Latent Alignment Loss for our proposed \gls{mdt} policy. 
    We analyse the impact of both auxiliary tasks on the ABCD CALVIN challenge. 
    The results show the average rollout length over 1000 instruction chains averaged over 3 seeds. }
    \label{fig: CALVIN SSL Loss Ablation}
\end{figure}
\begin{table*}[]
    \centering
    \begin{tabular}{l|l|cccccc|c}
    Goal Modality  & Model & Task 1 & Task 2 & Task 3 & Task 4 & Task 5 & Task 6 & Avg. Rollout Length  \\
        \midrule
  \multirow{3}{*}{Language}    & MT-ACT   & 50\% & 18.75\% & 12.50\% & 0\% & 0\% & 0\% & 0.81 $\pm$ 1.01 \\
     & MDT      & 75\% & 43.75\% & 18.75\% & 0\% & 0\% & 0\% & 1.38 $\pm$ 1.05 \\
     & MDT + $\mathcal{L}_{\text{MGF}}$ + $\mathcal{L}_{\text{CLA}}$ & 81.25\% & 56.25\% & 12.50\% & 6.25\% & 0\% & 0\% &  \textbf{1.56} $\pm$ \textbf{1.06} \\
     \midrule
 \multirow{3}{*}{Images}      & MT-ACT   & 12.50\% & 0\% & 0\% & 0\% & 0\% & 0\% & 0.13 $\pm$ 0.33 \\
     & MDT      & 12.50\% & 12.50\% & 6.25\% & 6.25\% & 0\% & 0\% & 0.38 $\pm$ 1.05 \\
     & MDT + $\mathcal{L}_{\text{MGF}}$ + $\mathcal{L}_{\text{CLA}}$ & 37.50\% & 6.25\% & 6.25\% & 6.25\% & 0\% & 0\% & \textbf{0.56} $\pm$ \textbf{1.00} \\
      \bottomrule
    \end{tabular}
    \caption{The average rollout length of the different approaches evaluated on the challenging long-horizon real robot play kitchen dataset. The performance is averaged over $4$ instruction chains with $4$ rollouts each. \gls{mdt} clearly outperforms MT-ACT. The performance of \gls{mdt} is further increased substantially by the auxiliary \gls{cla} and \gls{mgf} objectives. The relatively short avg rollout lengths emphasize how challenging this setting is, even for strong state-of-the-art methods such as MT-ACT.}
    \label{tab: real robot experiments}
\end{table*}

The promising results of \gls{mdt} compared to the state-of-the-art \gls{mtact} baseline in this challenging setting strongly suggests an affirmative answer to Question (III), especially with the proposed auxiliary \gls{cla} and \gls{mgf} objectives.

\subsection{Evaluation of Auxiliary Losses}\label{subsec: EvalAuxLosses}

Next, we investigate the significance of our auxiliary self-supervised loss functions, specifically the \gls{cla} and \gls{mgf} loss, on \gls{mdt}'s performance. 
Figure \ref{fig: CALVIN SSL Loss Ablation} shows the performance metrics of the ablated versions with and without these losses.
The inclusion of \gls{mgf} notably enhances \gls{mdt}'s performance on the CALVIN ABCD$\rightarrow$D benchmark, improving average rollout lengths by over $25\%$.
Detailed results supporting the essential role of these auxiliary tasks in \gls{mdt}-V are presented in Table~\ref{tab: MDTV ssl abalations} within the Appendix, showing that \gls{mdt}-V surpasses all baselines with an average rollout length of $4.12$ even in the absence of these two losses.

We further study the impact of \gls{mgf} and \gls{cla} on the LIBERO benchmark (summarized in Table \ref{tab: libero ablations}), where the auxiliary objectives improve \gls{mdt}'s success rates in $4$ out of $5$ task suites, achieving more than a $8.5\%$ increase on average. 
The results of these experiments are summarized in Table \ref{tab: libero ablations}.
Interestingly, we observe a synergistic effect when both losses are applied together. 
Overall, on LIBERO the performance impact of \gls{cla} is higher than that of \gls{mgf}, especially on the LIBERO-90 benchmark. 
A high number of different task labels, as it is the case for LIBERO-90, automatically implies a higher number of contrastive labels for any given tasks. 
We hypothesize that this increased number of contrastive labels leads to the higher performance impact of \gls{cla} for LIBERO-90.
The impact on the LIBERO-Goal is the lowest, given the high initial task success rate.
However, the LIBERO-Long benchmark does not seem to benefit from either \gls{mgf} or \gls{cla}. 
The demonstrations of the LIBERO-Long benchmark consist of several sub-tasks each with a single high-level description for the entire task.
We assume that this lack of sub-goals prevents the auxiliary losses from providing notable benefits.

To investigate if \gls{mgf} provides a generally beneficial auxiliary objective we integrate it with \gls{mtact} and evaluate the model for the full CALVIN ABCD$\rightarrow$D benchmark, as detailed in \autoref{tab: MT-ACT Ablations} in the Appendix.
\gls{mgf} significantly boosts \gls{mtact}'s average CALVIN performance by $44\%$, without any other modifications to the model or its hyperparameters. 
Similarly to the \gls{mdt} results, \gls{mgf} also enhances the performance of \gls{mtact} to learn better from multimodal goals with few language annotations. 
These positive outcomes for \gls{mdt}, along with its effective application to other transformer-based policies, positively answer research questions (IIa) and (IIb).

\begin{figure}
    \centering
    \includegraphics[width=\linewidth]{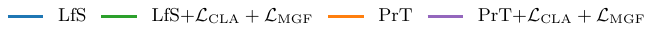}
    \includegraphics[width=\linewidth]{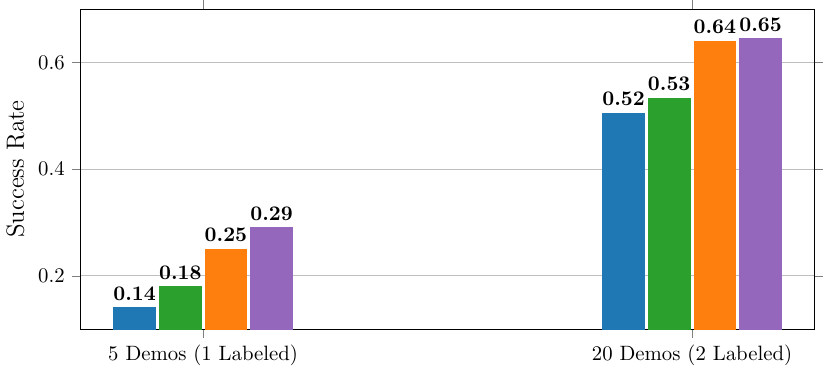}
    \caption{Study on the performance of our proposed \gls{mgf} and \gls{cla} objectives for pretraining on action-free data.  
    We pretrain MDT on LIBERO-90 with the objectives and test the average performance on all LIBERO-Long tasks with different number of demonstrations. 
    The results show the success rate averaged over 20 rollouts for all 10 tasks and 3 seeds. 
    LfS refers to trained from scratch and PrT are all pretrained models.}
    \label{fig: PreTraining LIBERO}
\end{figure}

\subsection{Additional Ablation Studies}

This section investigates several design choices for \gls{mdt} and the proposed auxiliary objectives. 
First, we explore the importance of using pre-trained CLIP embeddings and test \gls{mgf} and \gls{cla} as pretraining objective for diffusion policies. 
Next, we ablate several design decisions of \gls{mgf} and our proposed transformer architecture and how they impact performance. 

\textbf{Choice of Goal Image Encoder.}
We analyze the importance of using pre-trained CLIP encoders for \gls{mdt} to process goal images. 
\gls{mdt} is tested with ResNet Encoders for goal images, which are trained from scratch together with text embedding of a frozen CLIP model on CALVIN ABCD.
The \gls{mdt} model without our auxiliary tasks achieves an average rollout length of $3.34$, which is equal to the \gls{mdt} variant with the frozen CLIP embeddings.
The \gls{mdt} variant with both auxiliary objectives achieves an average rollout length of $4.31$, which is slightly worse compared to the variant with pre-trained encoders average performance of $4.41$. 
The detailed results of this experiment are summarized in Table \ref{tab: MDTV ssl abalations} of the Appendix. 
Overall, the ablation shows that pretrained CLIP image embeddings are not required for \gls{mdt} to succeed in learning from partially labeled datasets.


\textbf{Pretraining with \gls{mgf} and \gls{cla}.}
In this analysis, we assess the capability of \gls{cla} and \gls{mgf} for pretraining MDT on only video-data without access to the robot actions. 
Therefore, we pretrain \gls{mdt} with both auxiliary tasks on LIBERO-90 and test its downstream performance on LIBERO-Long, which contains unseen tasks.
We compare the pretrained variant against \gls{mdt} trained from scratch on this dataset and ablations, that omit auxiliary objectives for fine-tuning. 
For the LIBERO-Long experiment, we restrict all policies to learn from $5$ or $20$ demonstrations only.
For both settings we only label $10\%$ of all trajectories with the associated text instruction.
The results of these experiments are summarized in Figure \ref{fig: PreTraining LIBERO}.
Initializing \gls{mdt} with our pretrained weights boost the performance in the $5$ demonstration setting by $100\%$ and by $25\%$ in the variant with $20$ demonstrations.
Further, these experiments demonstrate, that both auxiliary objectives also improve performance of MDT with fewer demonstrations on LIBERO-Long independent on the pretraining.
The performance increase of MDT trained from scratch with both auxiliary objectives also verifies, that \gls{cla} and \gls{mgf} help \gls{mdt} to learn better in scenarios with small datasets.

\textbf{Masked Generative Foresight.}
Next, we study the different design choices of our \gls{mgf} loss and compare them against ablations.
Our primary focus is on assessing the impact of different masking ratios, ranging from $0.5$ to $1$, where $1$ corresponds to a full reconstruction of the initial future image. 
The results indicate that a masking ratio of $0.75$ achieves the best average performance, which is a value commonly used in other masking methods~\cite{karamcheti2023language}.
Thus, we use it as the default masking rate across all experiments in the paper.
Further details of this analysis are provided in Table~\ref{tab: masking rate MGF} in the Appendix.
Additionally, we investigate the ideal foresight distance for \gls{mgf} and evaluate it in two environments.
\gls{mdt} adopts a foresight distance of $v=3$ as this setting consistently delivers strong performance across various scenarios. 
While a higher foresight distance of $v=9$ does exhibit similar performance to a short distance of $v=1$, it is also associated with increased variance in results.
Further results of these investigations are presented in \autoref{tab: prediction horizon MGF} in the Appendix.

\textbf{Transformer Architecture.}
\gls{mdt} is tested against two Diffusion Transformer architectures previously described in~\cite{chi2023diffusionpolicy}.
The ablations are visualized in~\autoref{fig: transformer ablations} of the Appendix. 
These comparisons are conducted on the CALVIN ABCD$\rightarrow$D challenge, with detailed results featured in \ref{tab: MDTV ssl abalations} in the Appendix.
In the first ablation study, we incorporated a noise token as an additional input to the transformer encoder. 
This was done to assess the effect of excluding adaLN noise conditioning.
The second ablation represents the diffusion transformer architecture from~\cite{chi2023diffusionpolicy}, which does not use any encoder module.
\gls{mdt}-V, when trained without any auxiliary objective, achieves an average rollout length of $4.18$. 
The ablation without adaLN conditioning only achieves an average rollout length of $3.58$.
Notably, the complete omission of the transformer encoder led to a significantly lower average rollout length of $1.41$.
The experiments show, that the additional transformer encoder is crucial for diffusion policies to succeed in learning from different goals. 
In addition, separating the denoising process from the encoder and using adaLN conditioning further helps to boost performance and efficiency.

\subsection{Limitations}

While MDT shows strong performance on learning from multimodal goals, it still has several limitations: 
1) While we verify the effectiveness of our method in many tasks, \gls{mgf} and \gls{cla} do not increase the performance on LIBERO-Long,
2) The performance impact of \gls{mgf} and \gls{cla} varies across different benchmarks. On LIBERO, \gls{cla} has a higher impact, and on CALVIN, it's the opposite.
3) Diffusion Policies require multiple forward passes to generate an action sequence, resulting in lower inference speed compared to non-diffusion approaches.
4) The average rollout lengths of \gls{mdt} and the baseline on the real robot multi-tasks are relatively short. 
We credit this to the difficulty of the setting itself. Learning from partially labeled,  unsegmented, long-horizon play data is a very challenging task. We further hypothesize that the placement of the cameras is not ideal, as the robot significantly suffers from self-occlusion. The introduction on an in-hand camera could alleviate this problem.

\section{Conclusion}

In this work, we introduce \gls{mdt}, a novel continuous-time diffusion policy adept at learning long-horizon manipulation from play, requiring as little as $2\%$ language labels for effective training. 
To further improve effectiveness, we propose \gls{mgf} and \gls{cla} as simple, yet highly effective auxiliary objectives to learn more expressive behavior from multimodal goal specifications.
By reconstructing future states from multimodal goal specification and aligning these state representations in the latent space, the auxiliary objectives improve downstream policy learning without additional cost during inference. 
We rigorously tested \gls{mdt} across a diverse set of 184 tasks in both simulated environments and real-world settings. 
These extensive experiments not only validate our proposed auxiliary loss but also demonstrate the efficiency of the \gls{mdt} policy.
Notably, \gls{mdt} sets two records on the CALVIN benchmark and improves over prior sota with an average $15\%$ absolute performance increase.
Moreover, in detailed studies, we demonstrate that our auxiliary objectives improve learning from multimodal goals.

In future work, we would like to investigate the advantages of additional goal-modalities like sketches in \gls{mdt}. 
Furthermore, we plan to scale \gls{mdt} towards a versatile foundation policy by pretraining the model on the large-scale, partially labeled Open-X-Embodiment dataset~\cite{open_x_embodiment_rt_x_2023}.

\section*{Acknowledgments}

The work was funded by the German Research Foundation (DFG) – 448648559. 
The authors also acknowledge support by the state of Baden-Württemberg through HoreKa supercomputer funded by the Ministry of Science, Research and the
Arts Baden-Württemberg and by the German Federal Ministry of Education and Research.


\bibliographystyle{plainnat}
\bibliography{references}

\begin{thebibliography}{72}
\providecommand{\natexlab}[1]{#1}
\providecommand{\url}[1]{\texttt{#1}}
\expandafter\ifx\csname urlstyle\endcsname\relax
  \providecommand{\doi}[1]{doi: #1}\else
  \providecommand{\doi}{doi: \begingroup \urlstyle{rm}\Url}\fi

\bibitem[Alayrac et~al.(2022)Alayrac, Donahue, Luc, Miech, Barr, Hasson, Lenc, Mensch, Millican, Reynolds, et~al.]{alayrac2022flamingo}
Jean-Baptiste Alayrac, Jeff Donahue, Pauline Luc, Antoine Miech, Iain Barr, Yana Hasson, Karel Lenc, Arthur Mensch, Katherine Millican, Malcolm Reynolds, et~al.
\newblock Flamingo: a visual language model for few-shot learning.
\newblock \emph{Advances in Neural Information Processing Systems}, 35:\penalty0 23716--23736, 2022.

\bibitem[Becker et~al.(2023)Becker, Markgraf, Otto, and Neumann]{becker2023reinforcement}
Philipp Becker, Sebastian Markgraf, Fabian Otto, and Gerhard Neumann.
\newblock Reinforcement learning from multiple sensors via joint representations.
\newblock \emph{arXiv preprint arXiv:2302.05342}, 2023.

\bibitem[Bharadhwaj et~al.(2023)Bharadhwaj, Vakil, Sharma, Gupta, Tulsiani, and Kumar]{bharadhwaj2023roboagent}
Homanga Bharadhwaj, Jay Vakil, Mohit Sharma, Abhinav Gupta, Shubham Tulsiani, and Vikash Kumar.
\newblock Roboagent: Generalization and efficiency in robot manipulation via semantic augmentations and action chunking, 2023.

\bibitem[Blessing et~al.(2023)Blessing, Celik, Jia, Reuss, Li, Lioutikov, and Neumann]{blessing2023information}
Denis Blessing, Onur Celik, Xiaogang Jia, Moritz Reuss, Maximilian~Xiling Li, Rudolf Lioutikov, and Gerhard Neumann.
\newblock Information maximizing curriculum: A curriculum-based approach for learning versatile skills.
\newblock In \emph{Thirty-seventh Conference on Neural Information Processing Systems}, 2023.
\newblock URL \url{https://openreview.net/forum?id=7eW6NzSE4g}.

\bibitem[Chen et~al.(2023)Chen, Bahl, and Pathak]{chen2023playfusion}
Lili Chen, Shikhar Bahl, and Deepak Pathak.
\newblock Playfusion: Skill acquisition via diffusion from language-annotated play.
\newblock In \emph{7th Annual Conference on Robot Learning}, 2023.

\bibitem[Chi et~al.(2023)Chi, Feng, Du, Xu, Cousineau, Burchfiel, and Song]{chi2023diffusionpolicy}
Cheng Chi, Siyuan Feng, Yilun Du, Zhenjia Xu, Eric Cousineau, Benjamin Burchfiel, and Shuran Song.
\newblock Diffusion policy: Visuomotor policy learning via action diffusion.
\newblock In \emph{Proceedings of Robotics: Science and Systems (RSS)}, 2023.

\bibitem[Collaboration(2023)]{open_x_embodiment_rt_x_2023}
Open X-Embodiment Collaboration.
\newblock Open {X-E}mbodiment: Robotic learning datasets and {RT-X} models.
\newblock \url{https://arxiv.org/abs/2310.08864}, 2023.

\bibitem[Cui et~al.(2023)Cui, Wang, Shafiullah, and Pinto]{cui2023from}
Zichen~Jeff Cui, Yibin Wang, Nur Muhammad~Mahi Shafiullah, and Lerrel Pinto.
\newblock From play to policy: Conditional behavior generation from uncurated robot data.
\newblock In \emph{International Conference on Learning Representations}, 2023.
\newblock URL \url{https://openreview.net/forum?id=c7rM7F7jQjN}.

\bibitem[Driess et~al.(2023)Driess, Xia, Sajjadi, Lynch, Chowdhery, Ichter, Wahid, Tompson, Vuong, Yu, et~al.]{driess2023palm}
Danny Driess, Fei Xia, Mehdi~SM Sajjadi, Corey Lynch, Aakanksha Chowdhery, Brian Ichter, Ayzaan Wahid, Jonathan Tompson, Quan Vuong, Tianhe Yu, et~al.
\newblock Palm-e: An embodied multimodal language model.
\newblock \emph{arXiv preprint arXiv:2303.03378}, 2023.

\bibitem[Du et~al.(2023)Du, Yang, Dai, Dai, Nachum, Tenenbaum, Schuurmans, and Abbeel]{du2023learning}
Yilun Du, Mengjiao Yang, Bo~Dai, Hanjun Dai, Ofir Nachum, Joshua~B Tenenbaum, Dale Schuurmans, and Pieter Abbeel.
\newblock Learning universal policies via text-guided video generation.
\newblock \emph{arXiv preprint arXiv:2302.00111}, 2023.

\bibitem[Fu et~al.(2024)Fu, Zhao, and Finn]{fu2024mobile}
Zipeng Fu, Tony~Z Zhao, and Chelsea Finn.
\newblock Mobile aloha: Learning bimanual mobile manipulation with low-cost whole-body teleoperation.
\newblock \emph{arXiv preprint arXiv:2401.02117}, 2024.

\bibitem[Geng et~al.(2022)Geng, Liu, Lee, Schuurmans, Levine, and Abbeel]{geng2022multimodal}
Xinyang Geng, Hao Liu, Lisa Lee, Dale Schuurmans, Sergey Levine, and Pieter Abbeel.
\newblock Multimodal masked autoencoders learn transferable representations.
\newblock \emph{arXiv preprint arXiv:2205.14204}, 2022.

\bibitem[Gkanatsios et~al.(2023)Gkanatsios, Jain, Xian, Zhang, Atkeson, and Fragkiadaki]{gkanatsios2023energybased}
Nikolaos Gkanatsios, Ayush Jain, Zhou Xian, Yunchu Zhang, Christopher Atkeson, and Katerina Fragkiadaki.
\newblock {Energy-based Models are Zero-Shot Planners for Compositional Scene Rearrangement}.
\newblock In \emph{Robotics: Science and Systems}, 2023.

\bibitem[Grill et~al.(2020)Grill, Strub, Altch{\'e}, Tallec, Richemond, Buchatskaya, Doersch, Avila~Pires, Guo, Gheshlaghi~Azar, et~al.]{grill2020bootstrap}
Jean-Bastien Grill, Florian Strub, Florent Altch{\'e}, Corentin Tallec, Pierre Richemond, Elena Buchatskaya, Carl Doersch, Bernardo Avila~Pires, Zhaohan Guo, Mohammad Gheshlaghi~Azar, et~al.
\newblock Bootstrap your own latent-a new approach to self-supervised learning.
\newblock \emph{Advances in neural information processing systems}, 33:\penalty0 21271--21284, 2020.

\bibitem[Gu et~al.(2024)Gu, Kirmani, Wohlhart, Lu, Arenas, Rao, Yu, Fu, Gopalakrishnan, Xu, Sundaresan, Xu, Su, Hausman, Finn, Vuong, and Xiao]{gu2023rttrajectory}
Jiayuan Gu, Sean Kirmani, Paul Wohlhart, Yao Lu, Montserrat~Gonzalez Arenas, Kanishka Rao, Wenhao Yu, Chuyuan Fu, Keerthana Gopalakrishnan, Zhuo Xu, Priya Sundaresan, Peng Xu, Hao Su, Karol Hausman, Chelsea Finn, Quan Vuong, and Ted Xiao.
\newblock Rt-trajectory: Robotic task generalization via hindsight trajectory sketches.
\newblock In \emph{International Conference on Learning Representations}, 2024.

\bibitem[Ha et~al.(2023)Ha, Florence, and Song]{ha2023scaling}
Huy Ha, Pete Florence, and Shuran Song.
\newblock Scaling up and distilling down: Language-guided robot skill acquisition.
\newblock In \emph{7th Annual Conference on Robot Learning}, 2023.

\bibitem[He et~al.(2022)He, Chen, Xie, Li, Doll{\'a}r, and Girshick]{he2022masked}
Kaiming He, Xinlei Chen, Saining Xie, Yanghao Li, Piotr Doll{\'a}r, and Ross Girshick.
\newblock Masked autoencoders are scalable vision learners.
\newblock In \emph{Proceedings of the IEEE/CVF conference on computer vision and pattern recognition}, pages 16000--16009, 2022.

\bibitem[Ho et~al.(2020)Ho, Jain, and Abbeel]{ho2020denoising}
Jonathan Ho, Ajay Jain, and Pieter Abbeel.
\newblock Denoising diffusion probabilistic models.
\newblock \emph{Advances in Neural Information Processing Systems}, 33:\penalty0 6840--6851, 2020.

\bibitem[Jain et~al.(2022)Jain, Gkanatsios, Mediratta, and Fragkiadaki]{jain2022bottom}
Ayush Jain, Nikolaos Gkanatsios, Ishita Mediratta, and Katerina Fragkiadaki.
\newblock Bottom up top down detection transformers for language grounding in images and point clouds.
\newblock In \emph{European Conference on Computer Vision}, pages 417--433. Springer, 2022.

\bibitem[James et~al.(2020)James, Ma, Rovick~Arrojo, and Davison]{james2019rlbench}
Stephen James, Zicong Ma, David Rovick~Arrojo, and Andrew~J. Davison.
\newblock Rlbench: The robot learning benchmark \& learning environment.
\newblock \emph{IEEE Robotics and Automation Letters}, 2020.

\bibitem[Jiang et~al.(2023)Jiang, Gupta, Zhang, Wang, Dou, Chen, Fei-Fei, Anandkumar, Zhu, and Fan]{jiang2023vima}
Yunfan Jiang, Agrim Gupta, Zichen Zhang, Guanzhi Wang, Yongqiang Dou, Yanjun Chen, Li~Fei-Fei, Anima Anandkumar, Yuke Zhu, and Linxi Fan.
\newblock Vima: General robot manipulation with multimodal prompts.
\newblock In \emph{Fortieth International Conference on Machine Learning}, 2023.

\bibitem[Karamcheti et~al.(2023)Karamcheti, Nair, Chen, Kollar, Finn, Sadigh, and Liang]{karamcheti2023language}
Siddharth Karamcheti, Suraj Nair, Annie~S Chen, Thomas Kollar, Chelsea Finn, Dorsa Sadigh, and Percy Liang.
\newblock Language-driven representation learning for robotics.
\newblock \emph{arXiv preprint arXiv:2302.12766}, 2023.

\bibitem[Karras et~al.(2022)Karras, Aittala, Aila, and Laine]{karras2022elucidating}
Tero Karras, Miika Aittala, Timo Aila, and Samuli Laine.
\newblock Elucidating the design space of diffusion-based generative models.
\newblock In Alice~H. Oh, Alekh Agarwal, Danielle Belgrave, and Kyunghyun Cho, editors, \emph{Advances in Neural Information Processing Systems}, 2022.

\bibitem[Ko et~al.(2023)Ko, Mao, Du, Sun, and Tenenbaum]{Ko2023Learning}
Po-Chen Ko, Jiayuan Mao, Yilun Du, Shao-Hua Sun, and Joshua~B Tenenbaum.
\newblock {Learning to Act from Actionless Video through Dense Correspondences}.
\newblock \emph{arXiv:2310.08576}, 2023.

\bibitem[Laskin et~al.(2020)Laskin, Srinivas, and Abbeel]{laskin2020curl}
Michael Laskin, Aravind Srinivas, and Pieter Abbeel.
\newblock Curl: Contrastive unsupervised representations for reinforcement learning.
\newblock In \emph{International Conference on Machine Learning}, pages 5639--5650. PMLR, 2020.

\bibitem[Li et~al.(2023)Li, Belagali, Shang, and Ryoo]{li2023crossway}
Xiang Li, Varun Belagali, Jinghuan Shang, and Michael~S Ryoo.
\newblock Crossway diffusion: Improving diffusion-based visuomotor policy via self-supervised learning.
\newblock \emph{arXiv preprint arXiv:2307.01849}, 2023.

\bibitem[Li et~al.(2024)Li, Liu, Zhang, Yu, Xu, Wu, Cheang, Jing, Zhang, Liu, et~al.]{li2023vision}
Xinghang Li, Minghuan Liu, Hanbo Zhang, Cunjun Yu, Jie Xu, Hongtao Wu, Chilam Cheang, Ya~Jing, Weinan Zhang, Huaping Liu, et~al.
\newblock Vision-language foundation models as effective robot imitators.
\newblock In \emph{International Conference on Learning Representations}, 2024.

\bibitem[Lifshitz et~al.(2023)Lifshitz, Paster, Chan, Ba, and McIlraith]{lifshitz2023steve}
Shalev Lifshitz, Keiran Paster, Harris Chan, Jimmy Ba, and Sheila McIlraith.
\newblock Steve-1: A generative model for text-to-behavior in minecraft.
\newblock \emph{arXiv preprint arXiv:2306.00937}, 2023.

\bibitem[Liu et~al.(2023)Liu, Zhu, Gao, Feng, Liu, Zhu, and Stone]{liu2023libero}
Bo~Liu, Yifeng Zhu, Chongkai Gao, Yihao Feng, Qiang Liu, Yuke Zhu, and Peter Stone.
\newblock Libero: Benchmarking knowledge transfer for lifelong robot learning.
\newblock \emph{arXiv preprint arXiv:2306.03310}, 2023.

\bibitem[Lu et~al.(2022)Lu, Zhou, Bao, Chen, Li, and Zhu]{lu2022dpm}
Cheng Lu, Yuhao Zhou, Fan Bao, Jianfei Chen, Chongxuan Li, and Jun Zhu.
\newblock Dpm-solver++: Fast solver for guided sampling of diffusion probabilistic models.
\newblock \emph{arXiv preprint arXiv:2211.01095}, 2022.

\bibitem[Lynch and Sermanet(2020)]{lynch2020language}
Corey Lynch and Pierre Sermanet.
\newblock Language conditioned imitation learning over unstructured data.
\newblock \emph{arXiv preprint arXiv:2005.07648}, 2020.

\bibitem[Lynch et~al.(2020)Lynch, Khansari, Xiao, Kumar, Tompson, Levine, and Sermanet]{lynch2020learning}
Corey Lynch, Mohi Khansari, Ted Xiao, Vikash Kumar, Jonathan Tompson, Sergey Levine, and Pierre Sermanet.
\newblock Learning latent plans from play.
\newblock In \emph{Conference on robot learning}, pages 1113--1132. PMLR, 2020.

\bibitem[Lynch et~al.(2023)Lynch, Wahid, Tompson, Ding, Betker, Baruch, Armstrong, and Florence]{lynch2023interactive}
Corey Lynch, Ayzaan Wahid, Jonathan Tompson, Tianli Ding, James Betker, Robert Baruch, Travis Armstrong, and Pete Florence.
\newblock Interactive language: Talking to robots in real time.
\newblock \emph{IEEE Robotics and Automation Letters}, 2023.

\bibitem[Ma et~al.(2022)Ma, Sodhani, Jayaraman, Bastani, Kumar, and Zhang]{ma2022vip}
Yecheng~Jason Ma, Shagun Sodhani, Dinesh Jayaraman, Osbert Bastani, Vikash Kumar, and Amy Zhang.
\newblock Vip: Towards universal visual reward and representation via value-implicit pre-training.
\newblock In \emph{The Eleventh International Conference on Learning Representations}, 2022.

\bibitem[Majumdar et~al.(2023)Majumdar, Yadav, Arnaud, Ma, Chen, Silwal, Jain, Berges, Abbeel, Batra, Lin, Maksymets, Rajeswaran, and Meier]{majumdar2023where}
Arjun Majumdar, Karmesh Yadav, Sergio Arnaud, Yecheng~Jason Ma, Claire Chen, Sneha Silwal, Aryan Jain, Vincent-Pierre Berges, Pieter Abbeel, Dhruv Batra, Yixin Lin, Oleksandr Maksymets, Aravind Rajeswaran, and Franziska Meier.
\newblock Where are we in the search for an artificial visual cortex for embodied intelligence?
\newblock In \emph{Workshop on Reincarnating Reinforcement Learning at ICLR 2023}, 2023.
\newblock URL \url{https://openreview.net/forum?id=NJtSbIWmt2T}.

\bibitem[Mees et~al.(2022{\natexlab{a}})Mees, Hermann, and Burgard]{mees2022hulc}
Oier Mees, Lukas Hermann, and Wolfram Burgard.
\newblock What matters in language conditioned robotic imitation learning over unstructured data.
\newblock \emph{IEEE Robotics and Automation Letters (RA-L)}, 7\penalty0 (4):\penalty0 11205--11212, 2022{\natexlab{a}}.

\bibitem[Mees et~al.(2022{\natexlab{b}})Mees, Hermann, Rosete-Beas, and Burgard]{mees2022calvin}
Oier Mees, Lukas Hermann, Erick Rosete-Beas, and Wolfram Burgard.
\newblock Calvin: A benchmark for language-conditioned policy learning for long-horizon robot manipulation tasks.
\newblock \emph{IEEE Robotics and Automation Letters}, 2022{\natexlab{b}}.

\bibitem[Mees et~al.(2023)Mees, Borja-Diaz, and Burgard]{mees2023grounding}
Oier Mees, Jessica Borja-Diaz, and Wolfram Burgard.
\newblock Grounding language with visual affordances over unstructured data.
\newblock In \emph{2023 IEEE International Conference on Robotics and Automation (ICRA)}, pages 11576--11582. IEEE, 2023.

\bibitem[Myers et~al.(2023)Myers, He, Fang, Walke, Hansen-Estruch, Kolobov, Dragan, and Levine]{myers2023goal}
Vivek Myers, Andre~Wang He, Kuan Fang, Homer~Rich Walke, Philippe Hansen-Estruch, Andrey Kolobov, Anca Dragan, and Sergey Levine.
\newblock Goal representations for instruction following: A semi-supervised language interface to control.
\newblock In \emph{7th Annual Conference on Robot Learning}, 2023.

\bibitem[Nair et~al.(2022)Nair, Rajeswaran, Kumar, Finn, and Gupta]{nair2022r3m}
Suraj Nair, Aravind Rajeswaran, Vikash Kumar, Chelsea Finn, and Abhinav Gupta.
\newblock R3m: A universal visual representation for robot manipulation.
\newblock \emph{arXiv preprint arXiv:2203.12601}, 2022.

\bibitem[Pari et~al.(2021)Pari, Shafiullah, Arunachalam, and Pinto]{pari2021surprising}
Jyothish Pari, Nur~Muhammad Shafiullah, Sridhar~Pandian Arunachalam, and Lerrel Pinto.
\newblock The surprising effectiveness of representation learning for visual imitation, 2021.

\bibitem[Peebles and Xie(2023)]{peebles2023scalable}
William Peebles and Saining Xie.
\newblock Scalable diffusion models with transformers.
\newblock In \emph{Proceedings of the IEEE/CVF International Conference on Computer Vision}, pages 4195--4205, 2023.

\bibitem[Radford et~al.(2021)Radford, Kim, Hallacy, Ramesh, Goh, Agarwal, Sastry, Askell, Mishkin, Clark, et~al.]{radford2021learning}
Alec Radford, Jong~Wook Kim, Chris Hallacy, Aditya Ramesh, Gabriel Goh, Sandhini Agarwal, Girish Sastry, Amanda Askell, Pamela Mishkin, Jack Clark, et~al.
\newblock Learning transferable visual models from natural language supervision.
\newblock In \emph{International conference on machine learning}, pages 8748--8763. PMLR, 2021.

\bibitem[Radosavovic et~al.(2023)Radosavovic, Xiao, James, Abbeel, Malik, and Darrell]{radosavovic2023real}
Ilija Radosavovic, Tete Xiao, Stephen James, Pieter Abbeel, Jitendra Malik, and Trevor Darrell.
\newblock Real-world robot learning with masked visual pre-training.
\newblock In \emph{Conference on Robot Learning}, pages 416--426. PMLR, 2023.

\bibitem[Rana et~al.(2023{\natexlab{a}})Rana, Haviland, Garg, Abou-Chakra, Reid, and Suenderhauf]{rana2023sayplan}
Krishan Rana, Jesse Haviland, Sourav Garg, Jad Abou-Chakra, Ian Reid, and Niko Suenderhauf.
\newblock Sayplan: Grounding large language models using 3d scene graphs for scalable robot task planning.
\newblock In \emph{7th Annual Conference on Robot Learning}, 2023{\natexlab{a}}.
\newblock URL \url{https://openreview.net/forum?id=wMpOMO0Ss7a}.

\bibitem[Rana et~al.(2023{\natexlab{b}})Rana, Melnik, and S{\"u}nderhauf]{rana2023contrastive}
Krishan Rana, Andrew Melnik, and Niko S{\"u}nderhauf.
\newblock Contrastive language, action, and state pre-training for robot learning.
\newblock \emph{arXiv preprint arXiv:2304.10782}, 2023{\natexlab{b}}.

\bibitem[Reuss et~al.(2023)Reuss, Li, Jia, and Lioutikov]{reuss2023goal}
Moritz Reuss, Maximilian Li, Xiaogang Jia, and Rudolf Lioutikov.
\newblock Goal conditioned imitation learning using score-based diffusion policies.
\newblock In \emph{Proceedings of Robotics: Science and Systems (RSS)}, 2023.

\bibitem[Rosete-Beas et~al.(2022)Rosete-Beas, Mees, Kalweit, Boedecker, and Burgard]{rosete-beas2022latent}
Erick Rosete-Beas, Oier Mees, Gabriel Kalweit, Joschka Boedecker, and Wolfram Burgard.
\newblock Latent plans for task-agnostic offline reinforcement learning.
\newblock In \emph{6th Annual Conference on Robot Learning}, 2022.
\newblock URL \url{https://openreview.net/forum?id=ViYLaruFwN3}.

\bibitem[Scheikl et~al.(2023)Scheikl, Schreiber, Haas, Freymuth, Neumann, Lioutikov, and Mathis-Ullrich]{scheikl2023movement}
Paul~Maria Scheikl, Nicolas Schreiber, Christoph Haas, Niklas Freymuth, Gerhard Neumann, Rudolf Lioutikov, and Franziska Mathis-Ullrich.
\newblock Movement primitive diffusion: Learning gentle robotic manipulation of deformable objects.
\newblock \emph{arXiv preprint arXiv:2312.10008}, 2023.

\bibitem[Seo et~al.(2023)Seo, Kim, James, Lee, Shin, and Abbeel]{seo2023multi}
Younggyo Seo, Junsu Kim, Stephen James, Kimin Lee, Jinwoo Shin, and Pieter Abbeel.
\newblock Multi-view masked world models for visual robotic manipulation.
\newblock \emph{arXiv preprint arXiv:2302.02408}, 2023.

\bibitem[Shafiullah et~al.(2022)Shafiullah, Cui, Altanzaya, and Pinto]{shafiullah2022behavior}
Nur Muhammad~Mahi Shafiullah, Zichen~Jeff Cui, Ariuntuya Altanzaya, and Lerrel Pinto.
\newblock Behavior transformers: Cloning $k$ modes with one stone.
\newblock In \emph{Thirty-Sixth Conference on Neural Information Processing Systems}, 2022.
\newblock URL \url{https://openreview.net/forum?id=agTr-vRQsa}.

\bibitem[Shafiullah et~al.(2023)Shafiullah, Rai, Etukuru, Liu, Misra, Chintala, and Pinto]{shafiullah2023bringing}
Nur Muhammad~Mahi Shafiullah, Anant Rai, Haritheja Etukuru, Yiqian Liu, Ishan Misra, Soumith Chintala, and Lerrel Pinto.
\newblock On bringing robots home.
\newblock \emph{arXiv preprint arXiv:2311.16098}, 2023.

\bibitem[Shah et~al.(2023)Shah, Mart{\'\i}n-Mart{\'\i}n, and Zhu]{shah2023mutex}
Rutav Shah, Roberto Mart{\'\i}n-Mart{\'\i}n, and Yuke Zhu.
\newblock Mutex: Learning unified policies from multimodal task specifications.
\newblock In \emph{Conference on Robot Learning}, pages 2663--2682. PMLR, 2023.

\bibitem[Shridhar et~al.(2022{\natexlab{a}})Shridhar, Manuelli, and Fox]{shridhar2022cliport}
Mohit Shridhar, Lucas Manuelli, and Dieter Fox.
\newblock Cliport: What and where pathways for robotic manipulation.
\newblock In \emph{Conference on Robot Learning}, pages 894--906. PMLR, 2022{\natexlab{a}}.

\bibitem[Shridhar et~al.(2022{\natexlab{b}})Shridhar, Manuelli, and Fox]{shridhar2022peract}
Mohit Shridhar, Lucas Manuelli, and Dieter Fox.
\newblock Perceiver-actor: A multi-task transformer for robotic manipulation.
\newblock In \emph{Proceedings of the 6th Conference on Robot Learning (CoRL)}, 2022{\natexlab{b}}.

\bibitem[Shridhar et~al.(2023)Shridhar, Manuelli, and Fox]{shridhar2023perceiver}
Mohit Shridhar, Lucas Manuelli, and Dieter Fox.
\newblock Perceiver-actor: A multi-task transformer for robotic manipulation.
\newblock In \emph{Conference on Robot Learning}, pages 785--799. PMLR, 2023.

\bibitem[Song et~al.(2021)Song, Meng, and Ermon]{song2021denoising}
Jiaming Song, Chenlin Meng, and Stefano Ermon.
\newblock Denoising diffusion implicit models.
\newblock In \emph{ICLR}, 2021.

\bibitem[Song and Ermon(2019)]{song2019generative}
Yang Song and Stefano Ermon.
\newblock Generative modeling by estimating gradients of the data distribution.
\newblock \emph{Advances in Neural Information Processing Systems}, 32, 2019.

\bibitem[Song et~al.(2020)Song, Sohl-Dickstein, Kingma, Kumar, Ermon, and Poole]{song2020score}
Yang Song, Jascha Sohl-Dickstein, Diederik~P Kingma, Abhishek Kumar, Stefano Ermon, and Ben Poole.
\newblock Score-based generative modeling through stochastic differential equations.
\newblock In \emph{International Conference on Learning Representations}, 2020.

\bibitem[Song et~al.(2023)Song, Dhariwal, Chen, and Sutskever]{song2023consistency}
Yang Song, Prafulla Dhariwal, Mark Chen, and Ilya Sutskever.
\newblock Consistency models.
\newblock \emph{arXiv preprint arXiv:2303.01469}, 2023.

\bibitem[Stone et~al.(2023)Stone, Xiao, Lu, Gopalakrishnan, Lee, Vuong, Wohlhart, Kirmani, Zitkovich, Xia, Finn, and Hausman]{moo2023arxiv}
Austin Stone, Ted Xiao, Yao Lu, Keerthana Gopalakrishnan, Kuang-Huei Lee, Quan Vuong, Paul Wohlhart, Sean Kirmani, Brianna Zitkovich, Fei Xia, Chelsea Finn, and Karol Hausman.
\newblock Open-world object manipulation using pre-trained vision-language model.
\newblock In \emph{arXiv preprint}, 2023.

\bibitem[Vincent(2011)]{6795935}
Pascal Vincent.
\newblock A connection between score matching and denoising autoencoders.
\newblock \emph{Neural Computation}, 23\penalty0 (7):\penalty0 1661--1674, 2011.
\newblock \doi{10.1162/NECO_a_00142}.

\bibitem[Wang et~al.(2023)Wang, Fan, Sun, Zhang, Fei-Fei, Xu, Zhu, and Anandkumar]{wang2023mimicplay}
Chen Wang, Linxi Fan, Jiankai Sun, Ruohan Zhang, Li~Fei-Fei, Danfei Xu, Yuke Zhu, and Anima Anandkumar.
\newblock Mimicplay: Long-horizon imitation learning by watching human play.
\newblock In \emph{7th Annual Conference on Robot Learning}, 2023.
\newblock URL \url{https://openreview.net/forum?id=hRZ1YjDZmTo}.

\bibitem[Wu et~al.(2024)Wu, Jing, Cheang, Chen, Xu, Li, Liu, Li, and Kong]{wu2023unleashing}
Hongtao Wu, Ya~Jing, Chilam Cheang, Guangzeng Chen, Jiafeng Xu, Xinghang Li, Minghuan Liu, Hang Li, and Tao Kong.
\newblock Unleashing large-scale video generative pre-training for visual robot manipulation.
\newblock In \emph{International Conference on Learning Representations}, 2024.

\bibitem[Xian et~al.(2023)Xian, Gkanatsios, Gervet, and Fragkiadaki]{xian2023unifying}
Zhou Xian, Nikolaos Gkanatsios, Theophile Gervet, and Katerina Fragkiadaki.
\newblock Unifying diffusion models with action detection transformers for multi-task robotic manipulation.
\newblock In \emph{7th Annual Conference on Robot Learning}, 2023.

\bibitem[Xiao et~al.(2022{\natexlab{a}})Xiao, Chan, Sermanet, Wahid, Brohan, Hausman, Levine, and Tompson]{xiao2022robotic}
Ted Xiao, Harris Chan, Pierre Sermanet, Ayzaan Wahid, Anthony Brohan, Karol Hausman, Sergey Levine, and Jonathan Tompson.
\newblock Robotic skill acquisition via instruction augmentation with vision-language models.
\newblock \emph{arXiv preprint arXiv:2211.11736}, 2022{\natexlab{a}}.

\bibitem[Xiao et~al.(2022{\natexlab{b}})Xiao, Radosavovic, Darrell, and Malik]{xiao2022masked}
Tete Xiao, Ilija Radosavovic, Trevor Darrell, and Jitendra Malik.
\newblock Masked visual pre-training for motor control.
\newblock \emph{arXiv preprint arXiv:2203.06173}, 2022{\natexlab{b}}.

\bibitem[Zhan et~al.(2022)Zhan, Zhao, Pinto, Abbeel, and Laskin]{zhan2022learning}
Albert Zhan, Ruihan Zhao, Lerrel Pinto, Pieter Abbeel, and Michael Laskin.
\newblock Learning visual robotic control efficiently with contrastive pre-training and data augmentation.
\newblock In \emph{2022 IEEE/RSJ International Conference on Intelligent Robots and Systems (IROS)}, pages 4040--4047. IEEE, 2022.

\bibitem[Zhang et~al.(2024)Zhang, Lu, Wang, and Zhang]{zhang2022lad}
Edwin Zhang, Yujie Lu, William Wang, and Amy Zhang.
\newblock Language control diffusion: Efficiently scaling through space, time, and tasks.
\newblock In \emph{International Conference on Learning Representations}, 2024.

\bibitem[Zhao et~al.(2023)Zhao, Kumar, Levine, and Finn]{zhao2023learning}
Tony~Z Zhao, Vikash Kumar, Sergey Levine, and Chelsea Finn.
\newblock Learning fine-grained bimanual manipulation with low-cost hardware.
\newblock \emph{arXiv preprint arXiv:2304.13705}, 2023.

\bibitem[Zhou et~al.(2023)Zhou, Bing, Yao, Su, Yang, Huang, and Knoll]{zhou2023language}
Hongkuan Zhou, Zhenshan Bing, Xiangtong Yao, Xiaojie Su, Chenguang Yang, Kai Huang, and Alois Knoll.
\newblock Language-conditioned imitation learning with base skill priors under unstructured data.
\newblock \emph{arXiv preprint arXiv:2305.19075}, 2023.

\bibitem[Zitkovich et~al.(2023)Zitkovich, Yu, Xu, Xu, Xiao, Xia, Wu, Wohlhart, Welker, Wahid, Vuong, Vanhoucke, Tran, Soricut, Singh, Singh, Sermanet, Sanketi, Salazar, Ryoo, Reymann, Rao, Pertsch, Mordatch, Michalewski, Lu, Levine, Lee, Lee, Leal, Kuang, Kalashnikov, Julian, Joshi, Irpan, brian ichter, Hsu, Herzog, Hausman, Gopalakrishnan, Fu, Florence, Finn, Dubey, Driess, Ding, Choromanski, Chen, Chebotar, Carbajal, Brown, Brohan, Arenas, and Han]{zitkovich2023rt}
Brianna Zitkovich, Tianhe Yu, Sichun Xu, Peng Xu, Ted Xiao, Fei Xia, Jialin Wu, Paul Wohlhart, Stefan Welker, Ayzaan Wahid, Quan Vuong, Vincent Vanhoucke, Huong Tran, Radu Soricut, Anikait Singh, Jaspiar Singh, Pierre Sermanet, Pannag~R Sanketi, Grecia Salazar, Michael~S Ryoo, Krista Reymann, Kanishka Rao, Karl Pertsch, Igor Mordatch, Henryk Michalewski, Yao Lu, Sergey Levine, Lisa Lee, Tsang-Wei~Edward Lee, Isabel Leal, Yuheng Kuang, Dmitry Kalashnikov, Ryan Julian, Nikhil~J Joshi, Alex Irpan, brian ichter, Jasmine Hsu, Alexander Herzog, Karol Hausman, Keerthana Gopalakrishnan, Chuyuan Fu, Pete Florence, Chelsea Finn, Kumar~Avinava Dubey, Danny Driess, Tianli Ding, Krzysztof~Marcin Choromanski, Xi~Chen, Yevgen Chebotar, Justice Carbajal, Noah Brown, Anthony Brohan, Montserrat~Gonzalez Arenas, and Kehang Han.
\newblock {RT}-2: Vision-language-action models transfer web knowledge to robotic control.
\newblock In \emph{7th Annual Conference on Robot Learning}, 2023.
\newblock URL \url{https://openreview.net/forum?id=XMQgwiJ7KSX}.

\end{thebibliography}
\flushcolsend
\newpage
\appendix
\begin{figure*}[h]
    \centering
    \includegraphics[width=0.49\linewidth]{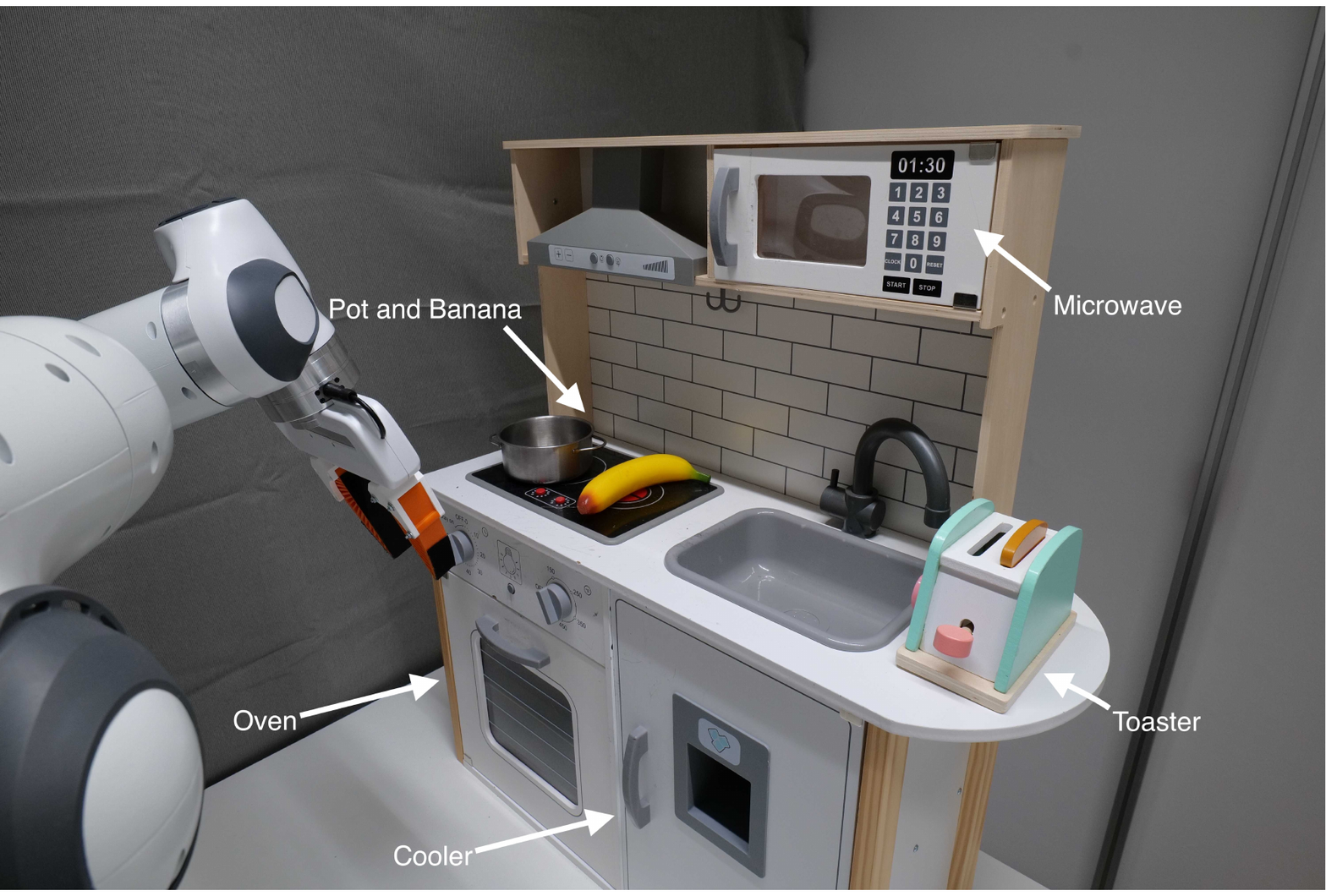}
    \hfill
    \includegraphics[width=0.49\linewidth]{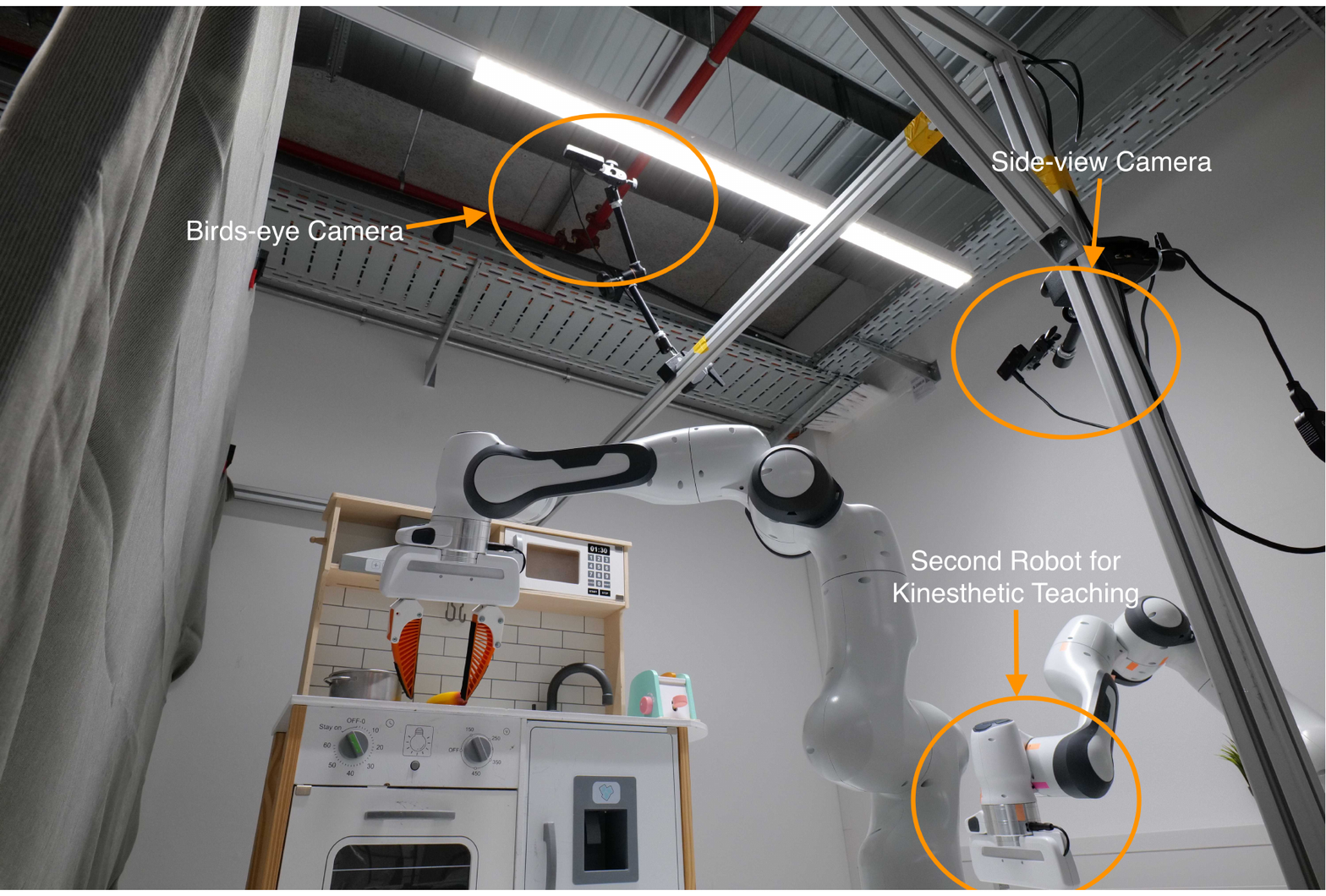}
    \caption{Overview of the real robot kitchens setup.
    The left image shows the play kitchen with all its objects, while the right image shows the cameras and second robot used for data collection with our robot used for teleoperation.}
    \label{fig:pdf-comparison}
\end{figure*}

\begin{figure*}[h]
    \centering
    \includegraphics[width=0.95\linewidth]{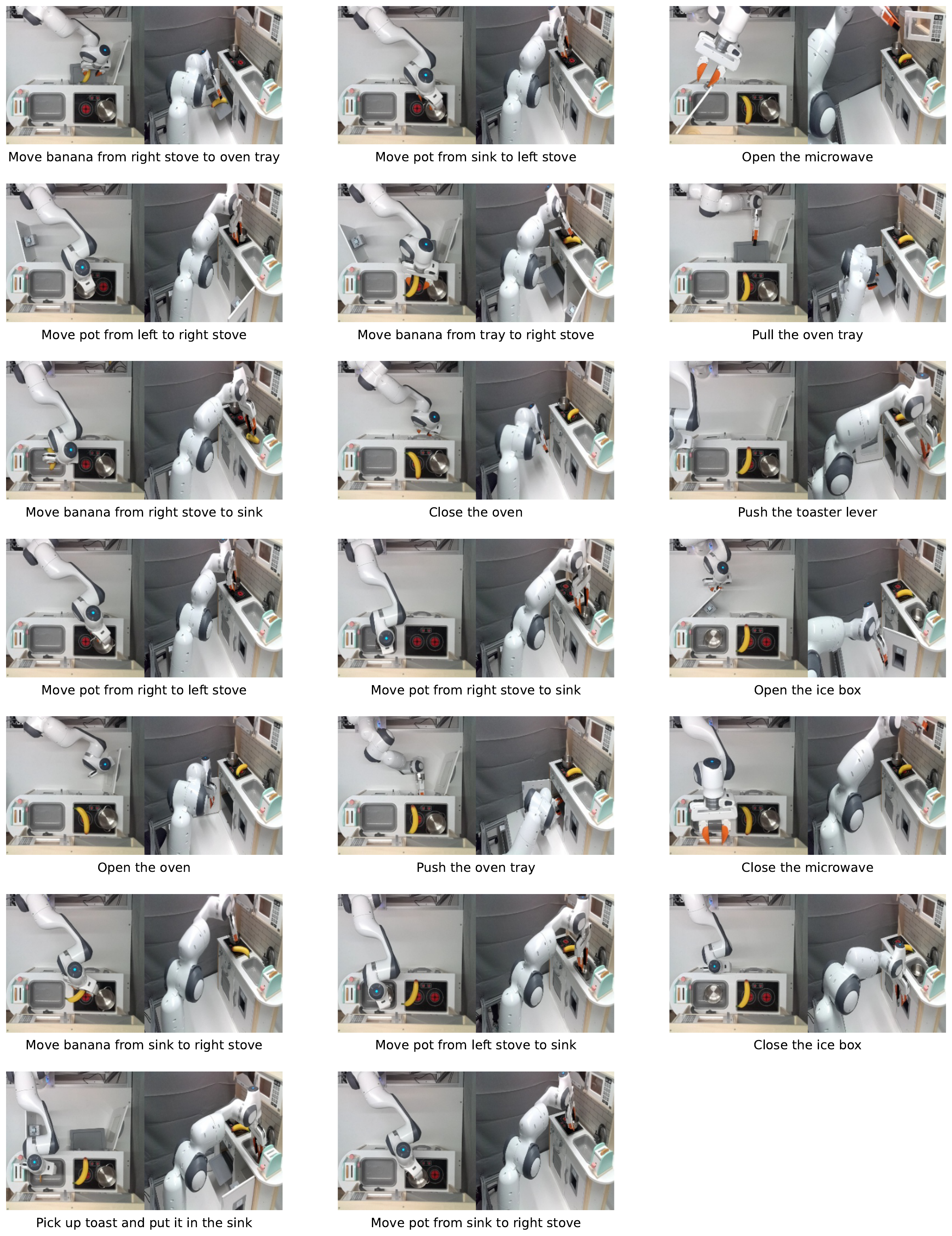}
    \caption{Overview of the 20 tasks recorded during play from the both camera perspectives. We test our policies on these tasks during evaluation after training on  a partially labeled dataset.}
    \label{fig:real-robot-tasks}
\end{figure*}

\begin{figure}[!htb]
    \centering
     \begin{algorithm}[H]  
            \caption{Score Matching Loss~\cite{song2019generative}}
            \label{alg:beso_training}
            \begin{algorithmic}[1]
                \State \textbf{Require:} Play Dataset $\mathcal{L}_{\text{play}}$
                \State \textbf{Require:} Score Model\newline $\mathbf{D}_{\theta}(\act, \state, \goal, \sigma_t)$
                \State \textbf{Require:} Noise Distribution:\newline $\text{LogLogistic}(\alpha, \beta)$
                \For{$i \in \{0, ...,N_{\text{train steps}} \}$ }
                    \State Sample $(\seq, \goal) \sim \mathcal{L}_{\text{play}}$ 
                    \State Sample $\sigma_t \sim \text{LogLogistic}(\alpha, \beta)$
                    \State Sample $\epsilon \sim \mathcal{N}(0, \sigma_t)$
                    \State $\mathcal{L}_{D_{\theta}} \gets \mathbb{E}_{\mathbf{\sigma}, \act, \boldsymbol{\epsilon}} \big[ \alpha (\sigma_t) \newline  \| D_{\theta}(\act + \boldsymbol{\epsilon}, \state, \goal, \sigma_t)  - \act  \|_2^2 \big] $
                \EndFor
            \end{algorithmic}
            \label{alg: diff training}
        \end{algorithm}
    \begin{minipage}{0.48\textwidth}
  \end{minipage}
    \begin{minipage}{0.48\textwidth}
\begin{algorithm}[H] 
\caption{DDIM Sampler as DPM-Solver-1~\cite{lu2022dpm, song2021denoising}}
\label{alg: diff inference}
    \begin{algorithmic}[1]
        \State \textbf{Require:} Current state $\state$, goal $\goal$
        \State \textbf{Require:} Score Model $D_{\theta}(\mathbf{a}, \state, \goal, \sigma)$
        \State \textbf{Require:} Noise scheduler $\sigma_i = \sigma(t_{i})$
        \State \textbf{Require:} Discrete time steps $t_{i \in \{ 0,.., N\}}$
        \State Draw sample $\mathbf{a}_{1:k,0} \sim \mathcal{N}( \mathbf{0}, \sigma_{0}^2\mathbf{I})$
        \For{$i \in \{0, ...,N-1 \}$ } 
            \State $t, t_{\text{next}} \gets t_{\text{fn}}(\sigma_i), t_{\text{fn}}(\sigma_{i+1})$
            \State $h \gets t_{\text{next}} - t$
            \State $\mathbf{a}_{1:k,i+1} \gets \frac{\sigma_{\text{fn}}(t_{\text{next}})}{ \sigma_{\text{fn}}(t)} \mathbf{a}_{1:k,i} \newline - \exp(-h)D_{\theta}(\mathbf{a}_{1:k,}, \state, \mathbf{g}, \sigma_{i})$
        \EndFor
        \State \textbf{return} $\mathbf{a}_{1:k,N}$
    \end{algorithmic}
\end{algorithm}
    \end{minipage}
\end{figure}

\subsection{Diffusion Model Training and Inference}
\label{sec: app score training and inference}

\begin{figure}
    \centering
    \includegraphics[width=\linewidth]{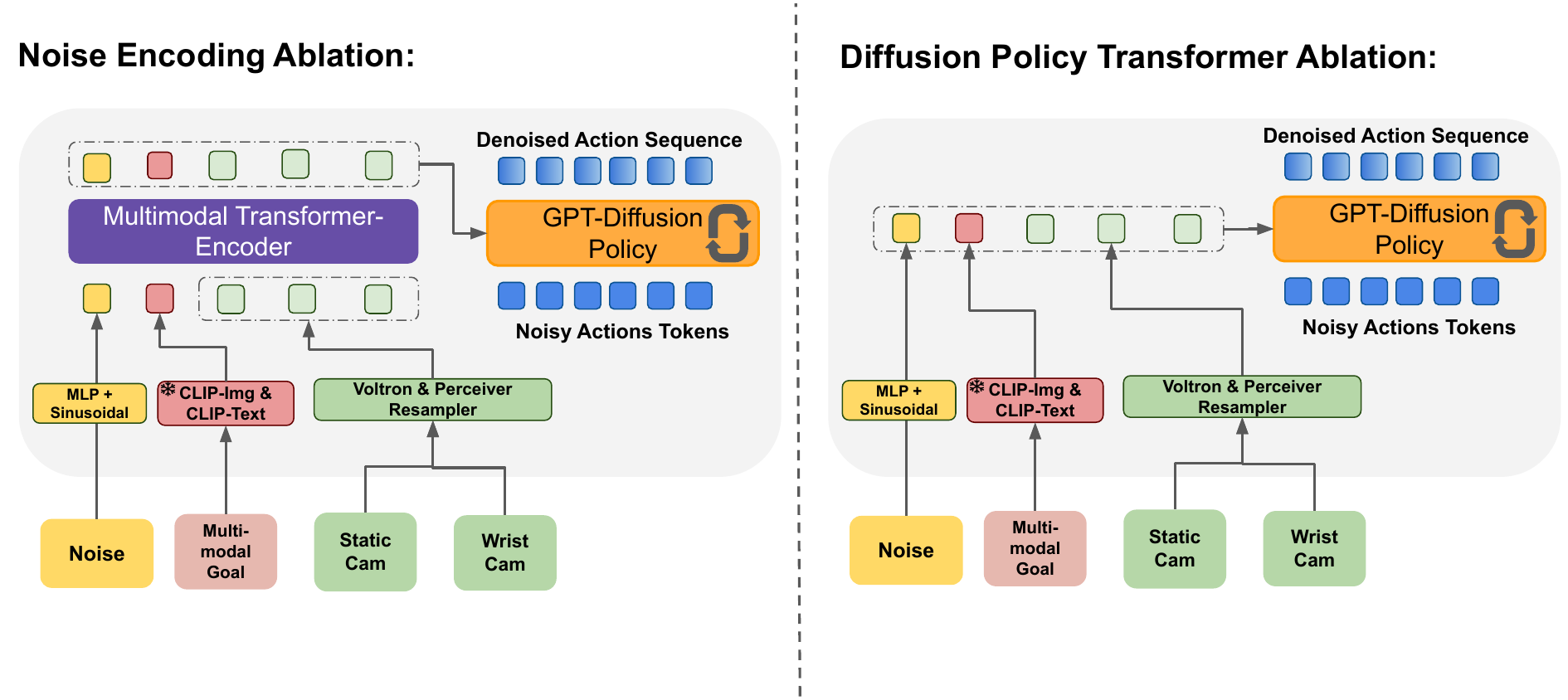}
    \caption{Overview of the two Diffusion Transformer Baseline Architectures used for the Ablation Study. 
    The first variant uses a transformer encoder but also processes the noise as a token. 
    The second one is the Transformer Diffusion policy presented in~\citet{chi2023diffusionpolicy} without any encoder. }
    \label{fig: transformer ablations}
\end{figure}

The training process for the score matching loss of MDT is summarized in Alg. \ref{alg: diff training} and the reverse diffusion process used for generating action chunks with DDIM sampler during the rollouts is summarized in Alg. \ref{alg: diff inference}.
Further, an overview of the used hyperparameters is given in Table \ref{tab:hyperparameters}.
To increase the performance, we deploy the preconditioning of \citet{karras2022elucidating}. 
This includes additional skip-connections and two pre-conditioning layers, which are conditioned on the current noise level $\sigma_t$ for effective balancing of the high range of noise levels from $0.001$ to $80$
\begin{equation}
\begin{split}
    D_{\theta}(\act| \state, \goal, \sigma_t) = c_{\text{skip}}(\sigma_t) \act\\+ c_{\text{out}}(\sigma_t)F_{\theta}(c_{\text{in}}(\sigma_t)\act, \state, \goal, c_{\text{noise}}(\sigma_t)).
\end{split}
\end{equation}
The utilized reconditioning functions are defined as: 
\begin{itemize}
    \item $c_{\text{skip}}= \sigma_{\text{data}}^2 / (\sigma_{\text{data}}^2 + \sigma_t^2)$
    \item $c_{\text{out}} = \sigma_t \sigma_{\text{data}} / \sqrt{\sigma_{\text{data}}^2 + \sigma_t^2}$
    \item $c_{\text{in}} = 1/ \sqrt{\sigma_{\text{data}}^2 + \sigma_t^2}$
    \item $ c_{\text{noise}} = 0.25 \ln (\sigma_t)$
\end{itemize}
They allow the model to decide, if it wants to predict the current noise, the denoised action sequence or something in between, depending on the current noise level \cite{karras2022elucidating}. 
For our noise distribution, we use a truncated Log-Logistic Distribution in range of $[\sigma_{\text{min}}, \sigma_{\text{max}} ]$.

\begin{table}[]
    \centering
    \begin{tabular}{c|ccc}
       \textbf{Hyperparameter}  & CALVIN  & LIBERO & Real World \\
       \midrule
      Number of Layers   & 6 & 2 & 2 \\
      Hidden Dimension & 192  & 192  & 192 \\
      Image resolution & 112 & 112 & 112 \\
      Masking Ratio & 0.75  & 0.75 & 0.75 \\
      MLP Ration & 4  & 4  & 4 \\
      Patch size & 16 & 16 & 16 \\
      Norm Pixel Loss & True & True & True \\
        \bottomrule
    \end{tabular}
    \caption{Overview of the chosen hyperparameters for our Image Demasking Model used in the \gls{mgf} loss, that uses a vision transformer architecture.}
    \label{tab: MGF model hyperparameters}
\end{table}

\begin{table}[]
    \centering
    \begin{tabular}{c|c}
       \textbf{Hyperparameter}  & Distill-D \\
       \midrule
      Action Chunk Size & 8 \\
      Timestep-embed Dimensions & 256 \\
      Image Encoder & ResNet18 \\
      Channel Dimensions & [512, 1024, 2048] \\
      Learning Rate & 1e-4 \\
      $\sigma_{\text{max}}$ &  80\\
        $\sigma_{\text{min}}$ & 0.001\\
        $\sigma_t$ & 0.5\\
        Time steps &  Exponential \\
        Sampler & DDIM \\
        Sampling Steps & 10 \\
        Trainable Parameters & 318 M \\
        Optimizer & AdamW  \\
        Betas & [0.9, 0.9]  \\
      Goal Image Encoder & CLIP ViT-B/16 \\
      Goal Lang Encoder & CLIP ViT-B/32 \\
      \bottomrule
    \end{tabular}
    \caption{Overview of the hyperparameters for Distill-D on the CALVIN and LIBERO benchmark. Our code is based on the Diffusion-policy implementation from~\citet{chi2023diffusionpolicy} with our continuous-time diffusion variant.
    To guarantee a fair comparison the hyperparameters for Distill-D and MDT regarding diffusion are the same.}
    \label{tab: Distill-D hyperparameters}
\end{table}

\begin{table*}[h]
\centering
\begin{tabular}{lccc}
\toprule
\textbf{Hyperparameter} &  MT-ACT & MDT-V & MDT \\
\hline
Number of Encoder Layers & 4 & 4 & 4 \\
Number of Decoder Layers & 6 & 4 & 6 \\
Attention Heads & 8 & 8 & 8\\
Action Chunk Size & 10 & 10 & 10\\
Goal Window Sampling Size & 49  & 49 & 49 \\
Hidden Dimension & 512 & 384 & 512 \\
Action Encoder Layers & 2 & - & - \\
Action Encoder Hidden Dim & 192 & - & - \\
Latent z dim & 32 & - & - \\
Image Encoder & ResNet18 & Voltron V-Cond & ResNet18 \\
Attention Dropout & 0.1 & 0.3 & 0.3 \\
Residual Dropout & 0.1 & 0.1 & 0.1\\
MLP Dropout & 0.1 & 0.05 & 0.05\\
Input Dropout & 0.0 & 0.0 & 0.0 \\
Optimizer & AdamW & AdamW & AdamW \\
Betas & [0.9, 0.9] & [0.9, 0.9] & [0.9, 0.9] \\
Transformer Weight Decay & 0.05 & 0.05 & 0.05 \\
Other weight decay & 0.05 & 0.05 & 0.05 \\
Batch Size & 512 & 512 & 512 \\
Trainable Parameters & 122 M & 40.0 M & 75.1 M \\
$\sigma_{\text{max}}$ & - & 80 & 80\\
$\sigma_{\text{min}}$ & - & 0.001 & 0.001\\
$\sigma_t$ & - & 0.5 & 0.5\\
Time steps & - & Exponential & Exponential \\
Sampler & - & DDIM & DDIM \\
Kl-$\beta$ & 50 & - & - \\
Contrastive Projection & - & MAP & Single Token 1 \\
Goal Image Encoder & CLIP ViT-B/16 & CLIP ViT-B/16 & CLIP ViT-B/16 \\
Goal Lang Encoder & CLIP ViT-B/32  & CLIP ViT-B/32 & CLIP ViT-B/32 \\
\bottomrule
\end{tabular}
\caption{Summary of all the Hyperparameters for the \gls{mdt} policy used in the CALVIN experiments and the ones of \gls{mtact}.}

\label{tab:hyperparameters}
\end{table*}

\begin{table}[]
    \centering
    \begin{tabular}{c|cc}
        Masking Rate  & CALVIN D & LIBERO-Spatial  \\
         \midrule
       0.5  &  3.7 $\pm$ 0.04 & 67.8 $\pm$ 0.3 \\
       0.75  & 3.72 $\pm$ 0.05 & 67.5 $\pm$ 0.2 \\
       1  & 3.68 $\pm$ 0.03 & 63.7 $\pm$ 0.3 \\
       \bottomrule
    \end{tabular}
    \caption{Ablation on different Masking Rates for Masked Generative Foresight, tested on CALVIN D$\rightarrow$D with \gls{mdt}-V and on LIBERO-Spatial with \gls{mdt}.}
    \label{tab: masking rate MGF}
\end{table}

\begin{table*}
\centering
\scalebox{1.25}{
\begin{tabular}{l|cc|ccccccc}
\toprule
\multirow{2}{*}{Method} & \multirow{2}{*}{$\mathcal{L}_{\text{CLA}}$} & \multirow{2}{*}{$\mathcal{L}_{\text{MGF}}$} & \multicolumn{6}{c}{No. Instructions in a Row (1000 chains)}  \\
  \cmidrule(lr){4-9}
 & & & 1 & 2 & 3 & 4 & 5 & \textbf{Avg. Len.} \\ 
 \midrule
MDT-V Abl. 1 & $\times$ & $\times$ & 0.914 & 0.782 & 0.675 & 0.588 & 0.487 & 3.58 $\pm$ 0.18 \\
MDT-V Abl. 2 & $\times$ & $\times$ & 0.693 & 0.405 & 0.190 & 0.092 & 0.031 & 1.41 $\pm$ 0.04 \\
 \midrule
MDT-V & $\times$ & $\times$ & 0.971 & 0.907 & 0.840 & 0.766 & 0.698 & 4.18 $\pm$ 0.10 \\
MDT-V & $\checkmark$ & $\times$ & 0.977 & 0.927 & 0.868 & 0.808 & 0.786 & 4.32 $\pm$ 0.06\\
MDT-V & $\times$ & $\checkmark$ & 0.986 & 0.946 & 0.903 & 0.851 & 0.794 & 4.48 $\pm$ 0.03\\
MDT-V & $\checkmark$ & $\checkmark$ & 0.989 & 0.958 & 0.916 & 0.862 & 0.801 & 4.52 $\pm$ 0.02\\
\midrule
MDT & $\times$ & $\times$ & 0.882 & 0.753 & 0.653 & 0.557 & 0.481 & 3.34 $\pm$ 0.06 \\

MDT+Goal ResNets  & $\times$ & $\times$ & 0.886 & 0.764 & 0.651 & 0.565 & 0.48 & 3.34 $\pm$ 0.05 \\

MDT & $\checkmark$ & $\checkmark$ & 0.978 & 0.938 & 0.888 & 0.831 & 0.77 & 4.41 $\pm$ 0.03 \\
MDT+Goal ResNets  & $\checkmark$ & $\checkmark$  & 0.978 & 0.923 & 0.862 & 0.807 & 0.735 & 4.31 $\pm$ 0.08 \\

\bottomrule
\end{tabular}}
\caption{Overview of the performance influence of MGF and Contrastive Alignment on MDT-V on the CALVIN ABCD$\rightarrow$D challenge. In addition, the performance of both transformer ablations are also shown. Moreover, results for MDT with ResNets as image goal encoders are reported with and without auxiliary objectives.
The results are reported over 1000 rollouts averaged over 3 seeds.}
\label{tab: MDTV ssl abalations}
\end{table*}

\subsection{Goal Sampling Strategy}

We experimented with different sampling strategies and ranges of future widows. 
We found that geometric sampling with a distribution probability of $p = 0.1$ works well in all tested settings on CALVIN. 
Other experiments with random sampling showed small drops in performance, while trials with key-state-based goal states similar to RLBench~\cite{james2019rlbench} did not work well in any setting. 
Thus, we decided to use the same strategy for CALVIN and real robot experiments.

\begin{table}[]
    \centering
    \begin{tabular}{c|cc}
         & CALVIN ABCD & LIBERO-Spatial  \\
         \midrule
       1  &  4.50 $\pm$ 0.02 & 64.4 $\pm$ 0.4 \\
       3  & \textbf{4.52} $\pm$ \textbf{0.02} & \textbf{67.5} $\pm$ \textbf{0.2} \\
       9  & 4.44 $\pm$ 0.03 & 65.6 $\pm$ 0.5 \\
       \bottomrule
    \end{tabular}
    \caption{Ablation on the best prediction horizon for Masked Generative Foresight, tested on CALVIN ABCD$\rightarrow$D with \gls{mdt}-V and LIBERO-Spatial with \gls{mdt}.}
    \label{tab: prediction horizon MGF}
\end{table}

\begin{table}[]
    \centering
    \begin{tabular}{l|c}
        Policy & Avg. Len. CALVIN  \\
         \midrule
        MT-ACT & 2.80 $\pm$ 0.03 \\
        MT-ACT + $\mathcal{L}_{\text{MGF}}$ & \textbf{4.03} $\pm$ \textbf{0.08} \\ 
    \bottomrule
    \end{tabular}
    \caption{Evaluation of the Performance Increase of the MT-ACT policy with the additional Masked Generative Foresight Loss on the CALVIN ABCD$\rightarrow$D challenge.}
    \label{tab: MT-ACT Ablations}
\end{table}

\subsection{Baseline Implementations}
\label{sec: baseline Implementations}
\textbf{MT-ACT.}
An overview of the used hyperparameters of MT-ACT is given in Table \ref{tab:hyperparameters}.
We tried to stay close to the recommended hyperparameters from the original paper \cite{bharadhwaj2023roboagent} but optimized the action prediction length and the Kl-$\beta$ factor for CALVIN.
Empirically, we found that FiLM-conditioned ResNets do not perform well conditioned on image-goal or in combination with not using any FiLM conditioning when having image goals. 
Thus, we adopted default ResNets as the vision encoders for MT-ACT, as other experiments with pre-trained Voltron-Encoders~\cite{karamcheti2023language} did not show good results and reduced the performance over $20\%$ on the CALVIN benchmark ABCD$\rightarrow$D. 

\textbf{Distill-D.} We use the reported hyperparameters of~\citet{chi2023diffusionpolicy} in combination with two ResNets18 and frozen CLIP encoders for visual and language goals as described in~\citet{ha2023scaling}. 
The resulting model contains 296.6 million trainable parameters in the 1D-CNN and an additional 22.4 million parameters for two ResNets-18. 
We also experimented with pretrained Voltron embeddings for Distill-D, however similar to MT-ACT Voltron did not show any performance improvements.

\section{Experiment Details}

\subsection{CALVIN Experiment Details}

For our experiments in the CALVIN benchmark, we use the evaluation protocol as described in~\citet{mees2022calvin} for consistent comparisons with other approaches from prior work~\cite{zhang2022lad, mees2022hulc, li2023vision}. 
All methods are trained with two images from the static camera and the wrist camera. 
We applied random shift augmentation for both images, then resized the static camera image to $224\times224$ pixels and the gripper camera images to $84\times84$ pixels. 
Finally, we normalized all images with the recommended values from CLIP. 
The action space is delta end-effector actions and gripper signals. 
While Distill-D has shown a preference for position-based control, our experiments across various models demonstrated superior performance with the default setting of velocity-based control.
We utilize the same CLIP image and language goal encoding models for all internally tested models.
For goal-image generation in the unlabeled data segment, we employ geometric sampling with a variance of $0.1$ and a future frame range of $20-50$ to randomly select goal images for our policies. 
All policies are trained for twenty thousand steps on the smaller dataset and thirty thousand on the complete dataset. 
Extended training duration did not yield performance enhancements, and considering the substantial computational demands of training on the full dataset, we avoided prolonged training duration.

\subsection{LIBERO Experiment Details}
\label{sec: libero}
The LIBERO task suites~\cite{liu2023libero} consists of $5$ different ones in the benchmark with 50 demonstrations per task. 
To emulate a scenario with sparse language labels, we divided the dataset into two segments: one set consists of single demonstrations accompanied by language annotations, and the other comprises 49 demonstrations without labels. 
For generating goal images, we utilized the final state of each rollout. 
We used the default end-effector action space in all our experiments. 
Consistent with the CALVIN setup, we employed identical image augmentation methods to prepare our data. 
We trained all models for $50$ epochs and then tested them on $20$ rollouts averaged over $3$ seeds. 
The benchmark is structured into five distinct task suites, each designed to test different aspects of robotic learning and manipulation:
\begin{itemize}
    \item \textbf{Spatial}: This suite emphasizes the robot's ability to understand and manipulate spatial relationships. Each task involves placing a bowl, among a constant set of objects, on a plate. 
    The challenge lies in distinguishing between two identical bowls that differ only in their spatial placement relative to other objects. 
    \item \textbf{Goal}: The Goal suite tests the robot's proficiency in understanding and executing varied task goals. 
    Despite using the same set of objects with fixed spatial relationships, each task in this suite differs in the ultimate goal, demanding that the robot continually adapt its motions and behaviors to meet these varying objectives.
    \item \textbf{Object}: Focused on object recognition and manipulation, this suite requires the robot to pick and place a unique object in each task. 
    \item \textbf{Long}: This suite comprises tasks that necessitate long-horizon planning and execution. 
    The Long suite is particularly challenging, as it tests the robot's ability to maintain performance and adaptability over extended task duration.
    \item \textbf{90}:  Offering a diverse set of 90 short-horizon tasks across five varied settings. 
\end{itemize}


\subsection{Real Robot Experiments}
\begin{table}[h]
\centering
\begin{tabular}{l|c|cccc}
\toprule
\textbf{Task} & \textbf{No.} & \textbf{MT-ACT} & \textbf{MDT} & \textbf{MDT MGF} \\ \hline
Banana from rt stove to sink      & 1   & 0   & 0.8 & 1.0  \\ 
Banana from sink to rt stove      & 2   & 0   & 0.8 & 1.0  \\ 
Pot from rt stove to sink         & 3   & 0   & 0   & 0    \\ 
Pot from sink to rt stove         & 4   & 0.4 & 0.8 & 0    \\ 
Pot from lt stove to sink         & 5   & 0   & 0.8 & 0.6  \\ 
Pot from sink to lt stove         & 6   & 0   & 0   & 1.0  \\ 
Pot from lt to rt stove           & 7   & 0.2 & 0.2 & 0    \\ 
Pot from rt to lt stove           & 8   & 0   & 0.6 & 1.0  \\ 
Open microwave                    & 9   & 1.0 & 1.0 & 1.0  \\ 
Close microwave                   & 10  & 0   & 1.0 & 0    \\ 
Open oven                         & 11  & 1.0 & 1.0 & 1.0  \\ 
Close oven                        & 12  & 0   & 0   & 0    \\
Open ice box                      & 13  & 1.0 & 1.0 & 1.0  \\ 
Close ice box                     & 14  & 0   & 1.0 & 0    \\ 
Pull oven tray                    & 15  & 0.6 & 0   & 0.2  \\ 
Push oven tray                    & 16  & 0.4 & 0   & 0.2  \\ 
Banana from rt stove to tray      & 17  & 0   & 0   & 0.4  \\ 
Banana from tray to rt stove      & 18  & 0   & 0   & 1.0  \\ 
Push toaster                      & 19  & 0   & 0.2 & 1.0  \\ 
Toast to sink                     & 20  & 0.4 & 1.0 & 1.0  \\ 
\bottomrule
\end{tabular}
\caption{Detailed results of our real robot kitchen experiments for single-task setting with language goals. All results are averaged over 5 rollouts. "right" and "left" are abbreviated with "rt" and "lt".}
\label{tab:single_task_real_world_experiments}
\end{table}

\begin{table}[h]
\centering
\tiny
\begin{tabular}{l|cccccc}
\toprule
\textbf{MT-ACT (Language Goals)} & \textbf{T1} & \textbf{T2} & \textbf{T3} & \textbf{T4} & \textbf{T5} & \textbf{T6} \\ \hline
1: Random    & 0.0 & 0.0 & 0.0 & 0.0 & 0.0 & 0.0 \\
2: Open Close All   & 1.0 & 0.75 & 0.5 & 0.0 & 0.0 & 0.0 \\
3: Stovetop Sink       & 0.5 & 0.0 & 0.0 & 0.0 & 0.0 & 0.0 \\
4: Oven        & 0.5 & 0.0 & 0.0 & 0.0 & 0.0 & -       \\
\bottomrule
\toprule
\textbf{MT-ACT (Image Goals)} & \textbf{T1} & \textbf{T2} & \textbf{T3} & \textbf{T4} & \textbf{T5} & \textbf{T6} \\ \hline
1: Random    & 0.0 & 0.0 & 0.0 & 0.0 & 0.0 & 0.0 \\
2: Open Close All   & 0.0 & 0.0 & 0.0 & 0.0 & 0.0 & 0.0 \\
3: Stovetop Sink       & 0.5 & 0.0 & 0.0 & 0.0 & 0.0 & 0.0 \\
4: Oven       & 0.0 & 0.0 & 0.0 & 0.0 & 0.0 & -       \\
\bottomrule

\toprule
\textbf{MDT (Language Goals)} & \textbf{T1} & \textbf{T2} & \textbf{T3} & \textbf{T4} & \textbf{T5} & \textbf{T6} \\ \hline
1: Random    & 0.75 & 0.75 & 0.0 & 0.0 & 0.0 & 0.0 \\
2: Open Close All   & 1.0 & 1.0 & 0.75 & 0.0 & 0.0 & 0.0 \\
3: Stovetop Sink       & 0.75 & 0.0 & 0.0 & 0.0 & 0.0 & 0.0 \\
4: Oven       & 0.5 & 0.0 & 0.0 & 0.0 & 0.0 & -       \\
\bottomrule
\toprule
\textbf{MDT (Image Goals)} & \textbf{T1} & \textbf{T2} & \textbf{T3} & \textbf{T4} & \textbf{T5} & \textbf{T6} \\ \hline
1: Random    & 0.5 & 0.5 & 0.25 & 0.25 & 0.0 & 0.0 \\
2: Open Close All   & 0.0 & 0.0 & 0.0 & 0.0 & 0.0 & 0.0 \\
3: Stovetop Sink       & 0.0 & 0.0 & 0.0 & 0.0 & 0.0 & 0.0 \\
4: Oven       & 0.0 & 0.0 & 0.0 & 0.0 & 0.0 & -       \\
\bottomrule

\toprule
\textbf{MDT + $\mathcal{L}_{\text{MGF}}$ + $\mathcal{L}_{\text{LCA}}$ (Language Goals)} & \textbf{T1} & \textbf{T2} & \textbf{T3} & \textbf{T4} & \textbf{T5} & \textbf{T6} \\ \hline
1. Random    & 1.0 & 0.75 & 0.25 & 0.25 & 0.0 & 0.0 \\
2. Open Close All   & 1.0 & 1.0 & 0.25 & 0.0 & 0.0 & 0.0 \\
3. Stovetop Sink       & 0.5 & 0.25 & 0.0 & 0.0 & 0.0 & 0.0 \\
4. Oven       & 0.75 & 0.25 & 0.0 & 0.0 & 0.0 & -       \\
\bottomrule
\toprule
\textbf{MDT + $\mathcal{L}_{\text{MGF}}$ + $\mathcal{L}_{\text{LCA}}$ (Image Goals)} & \textbf{T1} & \textbf{T2} & \textbf{T3} & \textbf{T4} & \textbf{T5} & \textbf{T6} \\ \hline
1: Random    & 0.25 & 0.25 & 0.25 & 0.25 & 0.0 & 0.0 \\ 
2: Open Close All   & 0.5 & 0.0 & 0.0 & 0.0 & 0.0 & 0.0 \\ 
3: Stovetop Sink       & 0.25 & 0.0 & 0.0 & 0.0 & 0.0 & 0.0 \\
4: Oven        & 0.5 & 0.0 & 0.0 & 0.0 & 0.0 & -       \\
\bottomrule
\end{tabular}
\caption{Detailed results of our long-horizon, real robot multi-task experiments.}
\label{tab:long-horizon-results}
\end{table}

\textbf{Detailed Environment Overview.}
Our real robot setup with the toy kitchen is visualized in \autoref{fig:pdf-comparison}. 
In the kitchen the robot can interact with the microwave positioned on the top right, the oven in the lower left half of the kitchen, the cooler on the lower-right side of the kitchen and the sink on the right side of the counter top. 
The robot is positioned next to a toy kitchen with the following additional objects: a banana, a pot, a toaster with toast. 
In total we create a set of $20$ diverse tasks for the  robot to learn from the partially labeled play data. 
All tasks of our dataset are shown in \autoref{fig:real-robot-tasks}. 

\textbf{Play Dataset Collection.}
Our volunteers collect play data with teleoperation with a leader and follower robot setup, which is visualized in \autoref{fig:pdf-comparison}. 
During the teleoperation, we collect the robot's proprioceptive sensor data and two images from our two static cameras with $6$ Hz. 
We extract the desired joint position as our action signal and normalize it in the range $[-1, 1]$.
To label the data with additional text instructions, we randomly sample short sequences from our uncurated play dataset and ask a human to describe the task in this segment. 
In total, we generate $360$ labeled short-horizon segments for the model training. 
For every task we query GPT-4 to generate different text instructions for more diverse language descriptions. 
We note, that training a single policy on such a small dataset of real world play data is very challenging for the models.

\textbf{Policy Training.}
We train all tested policies on our processed real robot data for around $24$ hours with a small cluster consisting of $4$ GPUs for around 100 epochs.
For checkpoint selection, we use the checkpoint with the lowest validation loss. 
For \gls{mtact}, we selected the last epoch since our prior experience with the benchmarks indicated an improvement in performance even when the validation loss began to rise again.

\textbf{Evaluation Details.}
We collect $10$ goal images of each task from our play dataset to test image-conditioning, and in addition we have a set of $10$ different text instructions for each task. 
Each policy is tested $5$ times from a starting position not seen in training with some added noise to it.
Further, we test our policies on long-horizon setup, where we define $4$ different instruction chains consisting of $5$ or $6$ tasks in sequence. 
During these rollouts, we observe the robot, if the policy completes the desired sub-task and only give it the next goal description if he manages to complete the prior task successfully. 
During our experiments, we further vary the orientation of the banana slightly for the robot to pick up, while we keep the toaster at the same position during all our experiments.
Detailed results for all our experiments for single task and multi-task setting are summarized in \autoref{tab:single_task_real_world_experiments} and \autoref{tab:long-horizon-results}. Task sequences used in the multi-task evaluation are listed in the following:
\begin{enumerate}
    \item \textbf{Random}: Push toaster, Pickup toast and put to sink, Banana from right stove to sink, Pot from left to right stove, Open oven, Open microwave
    \item \textbf{Open Close All}: Open microwave, Open oven, Open ice box, Close ice box, Close oven, Close microwave
    \item \textbf{Stovetop Sink}: Banana from right stove to sink, Push toaster, Pot from left to right stove, Pickup toast and put to sink, Pot from right to left stove, Banana from sink to right stove
    \item \textbf{Oven}: Open oven, Pull oven tray, Banana from right stove to oven tray, Push oven tray, Close oven
\end{enumerate}

\textbf{Failure Cases.}
Both variants of MDT struggle to solve all tasks related to moving the banana and the pot to specific positions. 
Especially the two tasks "Move the pot/banana from the right stove to the sink" is often misunderstood by all tested policies. 
We hypothesize, that the policies don't have enough labels to learn to differentiate between these similar states. Policies also struggle solving tasks where they need to close a door, or push the oven tray. These tasks are "Close oven/microwave/ice box" and "Push oven tray". Most of the time, the policies failed these tasks by a few millimeters. We hypothesize, that these tasks included tight actions which have varying degrees of "openness" and had quick demonstrations in the dataset. These factors combined is likely the reason why the policies failed these tasks by a narrow margin.

\begin{figure*}
    \centering
    \includegraphics[width=\linewidth]{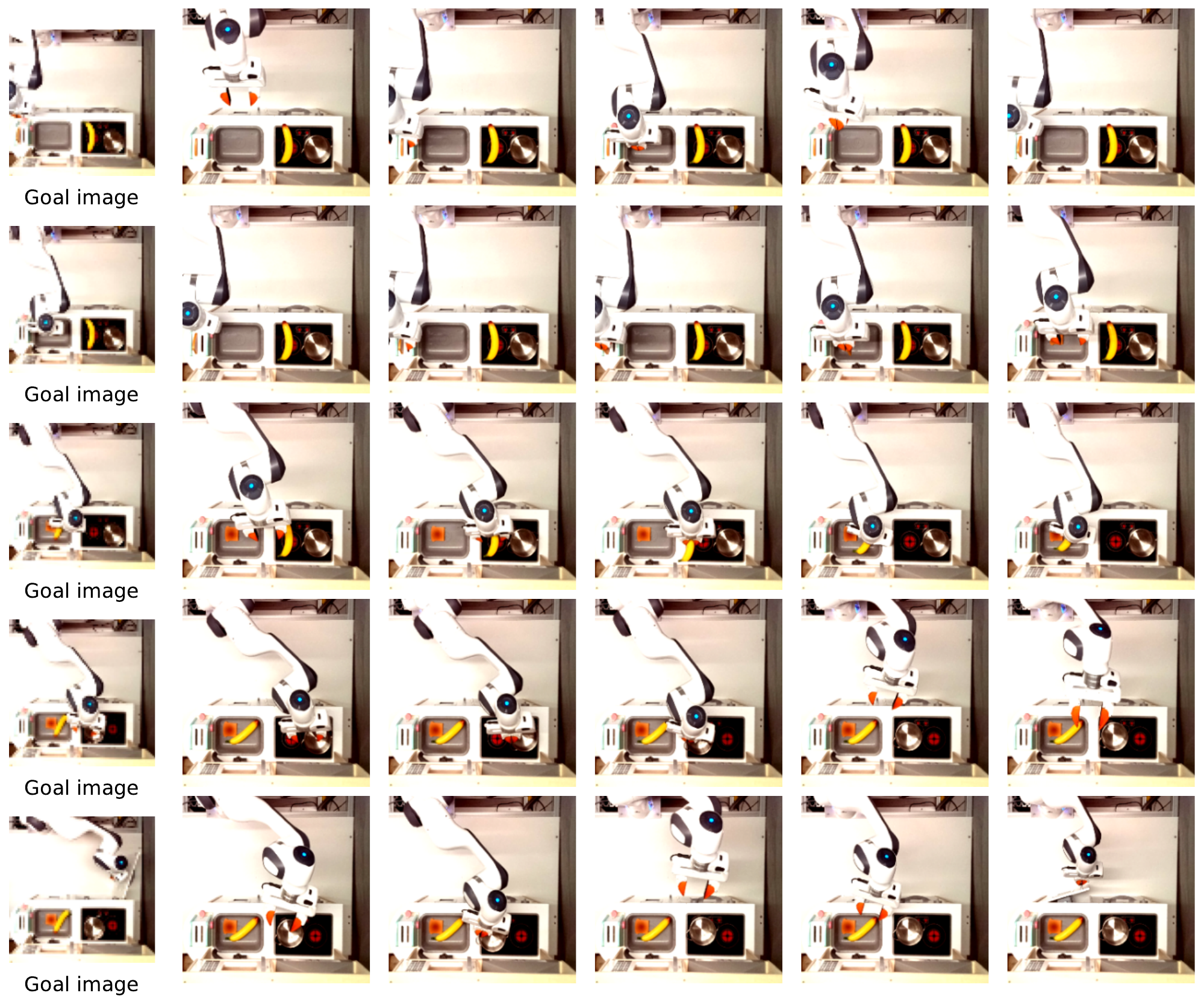}
    \caption{Real Robot rollouts with goal image conditioning. The first column shows the goal image used for the rollout. 4 out of 6 tasks are successful. 
    The robot fails to open the oven door and opens the ice box instead. }
    \label{fig: real robot rollouts}
\end{figure*}

\end{document}